%% file: main.tex
\pdfoutput=1

\documentclass[11pt]{article}

\usepackage[review]{EMNLP2023}

\usepackage{times}
\usepackage{latexsym}

\usepackage[T1]{fontenc}

\usepackage[utf8]{inputenc}

\usepackage{microtype}

\usepackage{inconsolata}
\usepackage{textcomp}

\usepackage{amsmath}
\usepackage{enumitem}
\usepackage{tcolorbox}

\usepackage{xcolor}
\usepackage{ifthen}
\usepackage[normalem]{ulem}

\definecolor{myorange}{HTML}{FEAE03}
\definecolor{myturquois}{HTML}{01AB8F}
\definecolor{mypink}{HTML}{D31876}

\definecolor{brightred}{HTML}{E55347} 
\definecolor{orange}{HTML}{FF8C00} 
\definecolor{yellowgreen}{HTML}{6B8E23} 
\definecolor{green}{HTML}{228B22} 
\newtcolorbox{bluebox}[1][]{
	float,
  	title=#1,
	colback=myturquois!4,
	colframe=myturquois,
        top=1pt,           
        bottom=1pt,        
        left=0pt,          
        right=0pt,          
        before skip=0.65em, after skip=0.75em,
}

\definecolor{darkgrey}{rgb}{0.53,0.53,0.53}
\definecolor{mygrey}{rgb}{0.9,0.9,0.9}
\definecolor{color1}{HTML}{006EB8}

\usepackage{tikz}
\usepackage[edges]{forest}
\usetikzlibrary{shapes, arrows.meta, positioning}
\usepackage{graphicx}
\usepackage{forest}
\usetikzlibrary{trees,positioning,shapes,shadows,arrows.meta}

\definecolor{myblue}{RGB}{159, 192, 230}
\definecolor{myblueline}{RGB}{87, 127, 185}
\definecolor{bluelight1}{RGB}{185, 211, 237}
\definecolor{bluelight2}{RGB}{213, 222, 239}
\definecolor{mygreen}{RGB}{168, 209, 201}
\definecolor{greenlight}{RGB}{220, 235, 234}
\definecolor{hidden-draw}{RGB}{177, 177, 177}
\definecolor{mygray}{RGB}{185, 185, 185}

\definecolor{lightcoral}{rgb}{0.94, 0.5, 0.5}
\definecolor{lightgreen}{rgb}{0.56, 0.93, 0.56}
\definecolor{harvestgold}{rgb}{0.98, 0.85, 0.40}
\definecolor{brightlavender}{rgb}{0.75, 0.58, 0.89}
\definecolor{capri}{rgb}{0.0, 0.75, 1.0}
\definecolor{carminepink}{rgb}{0.92, 0.3, 0.26}
\definecolor{celadon}{rgb}{0.67, 0.88, 0.69}
\definecolor{darkpastelgreen}{rgb}{0.01, 0.75, 0.24}

\newcommand{\commentisunseen}{0}

\newcommand{\commentp}[1]{
}

\newcommand{\fc}[1]{
\ifthenelse{\equal{\commentisunseen}{0}}{
{\color{blue}#1}}
{#1}
}
\newcommand{\FC}[1]{
\ifthenelse{\equal{\commentisunseen}{0}}{
{\color{blue}FC: #1}}
{}
}

\title{Knowledge Mechanisms in Large Language Models:\\ A Survey and Perspective}

\author{First Author \\b
  Affiliation / Address line 1 \\
  Affiliation / Address line 2 \\
  Affiliation / Address line 3 \\
  \texttt{email@domain} \\\And
  Second Author \\
  Affiliation / Address line 1 \\
  Affiliation / Address line 2 \\
  Affiliation / Address line 3 \\
  \texttt{email@domain} \\}

\author{
Mengru Wang\textsuperscript{1}\thanks{~~Equal Contribution.}, 
Yunzhi Yao\textsuperscript{1}\footnotemark[1],
Ziwen Xu\textsuperscript{1},
Shuofei Qiao\textsuperscript{1},
Shumin Deng\textsuperscript{2},\\
\textbf{Peng Wang\textsuperscript{1}, 
Xiang Chen\textsuperscript{1}, 
Jia-Chen Gu\textsuperscript{3}, 
Yong Jiang\textsuperscript{4}, 
Pengjun Xie\textsuperscript{4},} \\
\textbf{Fei Huang\textsuperscript{4},
Huajun Chen\textsuperscript{1},
Ningyu Zhang\textsuperscript{1}\thanks{~~Corresponding Author.}}\\
\textsuperscript{1}Zhejiang University,
~\textsuperscript{2}National University of Singapore, NUS-NCS Joint Lab, Singapore,\\ ~\textsuperscript{3}University of California, Los Angeles,~ \textsuperscript{4}Alibaba Group\\
\texttt{\{mengruwg,zhangningyu\}@zju.edu.cn}\\
}

\begin{document}

\maketitle
\begin{abstract}
Understanding knowledge mechanisms in Large Language Models (LLMs) is crucial for advancing towards trustworthy AGI. This paper reviews knowledge mechanism analysis from a novel taxonomy including knowledge utilization and evolution. Knowledge utilization delves into the mechanism of memorization, comprehension and application, and creation. Knowledge evolution focuses on the dynamic progression of knowledge within individual and group LLMs. Moreover, we discuss what knowledge LLMs have learned, the reasons for the fragility of parametric knowledge, and the potential dark knowledge (hypothesis) that will be challenging to address. We hope this work can help understand knowledge in LLMs and provide insights for future research.
\end{abstract}


\input{section/Introduction}

\input{section/Preliminary}

\input{section/Knowledge_Representation}

\input{section/Evolution}

\input{section/Application}

\input{section/Discussion}

\input{section/Future_Directions}

\input{section/Conclusion}

\section*{Limitations}
This work has some limitations as follows:

\paragraph{Hypothesis}
Despite reviewing a large body of literature and proposing several promising hypotheses, there are still some limitations.
On the one hand, there may be other hypotheses for knowledge utilization and evolution in LLMs. 
On the other hand, the accuracy of these hypotheses requires further exploration and validation over time.

\paragraph{Knowledge} There are various forms of knowledge representation.
However, due to current research constraints, this paper does not delve into space \cite{DBLP:journals/corr/abs-2403-00813}, time \cite{SpaceandTime}, event-based knowledge, and geoscience \cite{DBLP:journals/corr/abs-2401-00434}.

\paragraph{Reference} 
The field of knowledge mechanisms is developing rapidly and this paper may miss some important references.
Additionally, due to the page limit, we have omit certain technical details.
We will continue to pay attention to and supplement new works.

\paragraph{Models} 
Despite mentioning artificial neural models in this paper, knowledge mechanism analysis focuses on LLMs. We will continue to pay attention to other modal models progresses. Besides, all existing work has not considered models larger than 100 billion parameters. 
Whether the knowledge mechanisms within large-scale models are consistent with smaller ones remains to be studied.

\section*{Ethics Statement}
We anticipate no ethical or societal implications arising from our research.
However, we acknowledge that the internal mechanisms of large language models might be exploited for malicious purposes. 
We believe such malicious applications can be prevented through model access and legislative regulation.
More critically, a transparent model contributes to the development of safer and more reliable general artificial intelligence.

\section*{Acknowledgements}

We would like to express gratitude to the anonymous reviewers for their kind comments. 
This work was supported by the National Natural Science Foundation of China (No. 62206246, No. NSFCU23B2055, No. NSFCU19B2027), the Fundamental Research Funds for the Central Universities (226-2023-00138), Zhejiang Provincial Natural Science Foundation of China (No. LGG22F030011), Yongjiang Talent Introduction Programme (2021A-156-G), Information Technology Center and State Key Lab of CAD\&CG, Zhejiang University, and NUS-NCS Joint Laboratory (A-0008542-00-00).

\bibliography{anthology,custom}
\bibliographystyle{acl_natbib}

\input{section/appendix}

\end{document}

%% file: section/Introduction.tex
\section{Introduction}
\label{Introduction}

Knowledge is the cornerstone of intelligence and the continuation of civilization, furnishing us with foundational principles and guidance for navigating complex problems and emerging challenges \cite{DBLP:journals/aim/DavisSS93,DBLP:conf/wsdm/Choi22}.
Throughout the extensive history of evolution, we have dedicated our lives to cultivating more advanced intelligence by utilizing acquired knowledge and exploring the frontiers of unknown knowledge \cite{DBLP:books/daglib/0067119,DBLP:journals/cacm/HanZL21}.

As we know, Large language models (LLMs) are renowned for encapsulating extensive parametric knowledge \cite{DBLP:conf/emnlp/RobertsRS20,DBLP:conf/emnlp/SungLYJKK21,DBLP:conf/acl/CaoLHSYLXX20,DBLP:conf/naacl/ZhongFC21,Long-TailKnowledge,DBLP:journals/corr/abs-2008-09036,DBLP:conf/emnlp/PetroniRRLBWM19,DBLP:conf/acl/QiaoO0CYDTHC23,DBLP:conf/kbclm/KritharoulaLS23,DBLP:journals/corr/abs-2402-14273}, achieving unprecedented progress in application.
However, \textit{the knowledge mechanisms in LLMs for learning, storage, utilization, and evolution still remain mysterious} \cite{phillips2021four,DBLP:journals/corr/abs-2312-09230,DBLP:journals/corr/abs-2410-21815}.
Extensive works aim to demystify various types of knowledge in LLMs through knowledge neurons \cite{KN,chen2024analyzing} and circuits \cite{elhage2021mathematical,knowledgecircuit,CircuitBreakers}, yet these efforts, scattered across various tasks, await comprehensive review and analysis.

\begin{figure}
    \centering
    \includegraphics[width=0.5\textwidth]{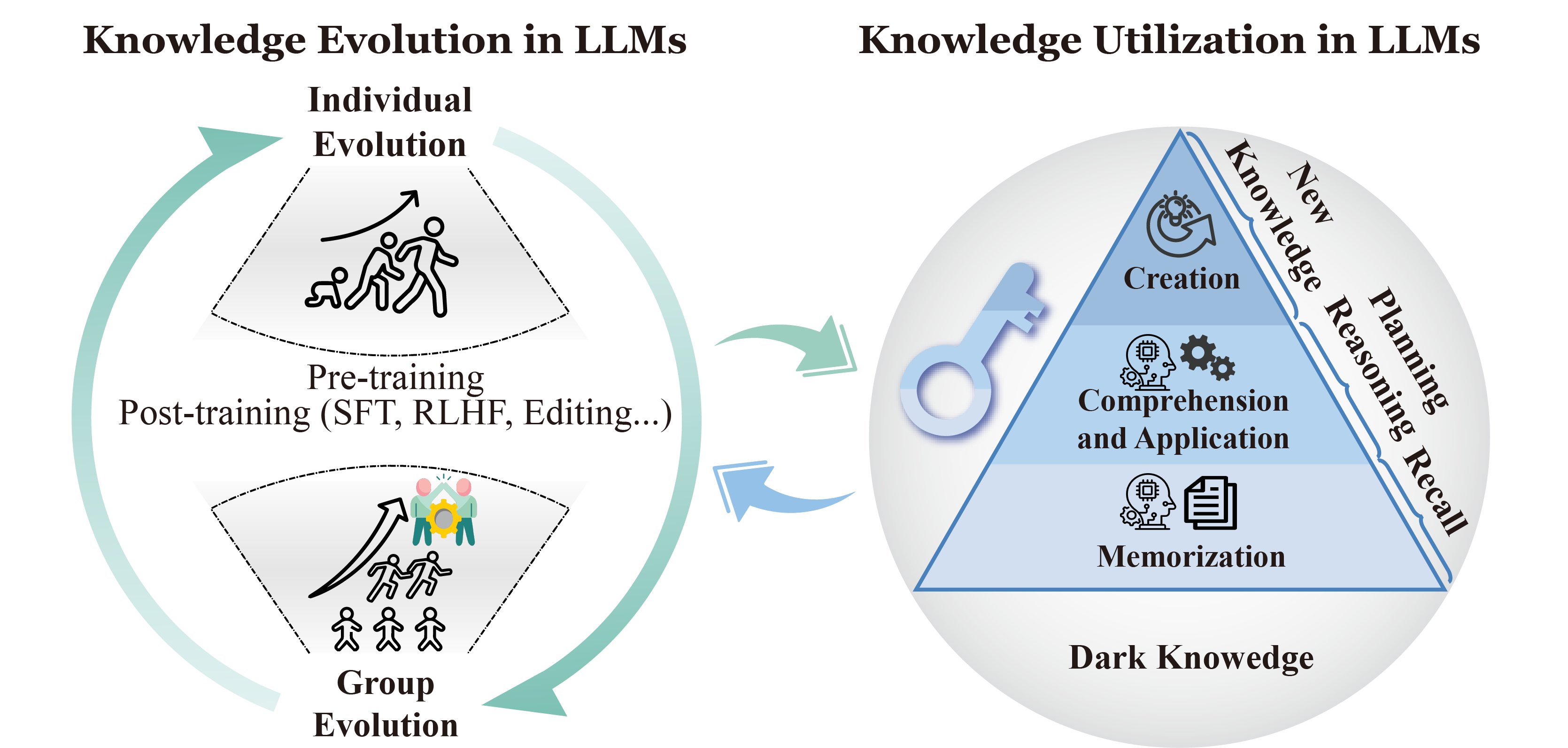}
    \caption{The analysis framework of knowledge mechanism within neural models includes knowledge evolution and utilization. 
    Dark knowledge denotes knowledge unknown to human or model (machine).
    We investigate the mechanisms of knowledge utilization (right) in LLMs during a specific period of their evolution (left). The knowledge limitations identified through mechanisms analysis will inspire subsequent evolution (left).}
    \label{fig:framwork}
    \vskip -0.1in
\end{figure}

As shown in Fig \ref{fig:framwork}, \textbf{this paper pioneeringly reviews the mechanism across the whole knowledge life cycle.}
We also propose a novel taxonomy for knowledge mechanisms in LLMs,  as illustrated in Fig \ref{taxonomy}, which encompasses knowledge utilization at a specific time and knowledge evolution across all periods of LLMs \footnote{Knowledge utilization focuses on \textit{static} knowledge at a specific period, while knowledge evolution explores the long-term \textit{dynamic} development of knowledge across individual and group LLMs.}.
Specifically, we introduce preliminaries of this field (\S \ref{Preliminary}) and review the knowledge utilization mechanism from a new perspective (\S \ref{Knowledge in LLMs}),
delve into the fundamental principles for knowledge evolution (\S \ref{Knowledge Evolution}).
Then, we investigate how to construct more efficient and trustworthy LLMs from the perspective of knowledge mechanism (\S \ref{Application}).
Later, We discuss open questions about the knowledge LLMs have and have not acquired (\S \ref{Discussion}).
Finally, we also provide some future directions (\S \ref{Future_Directions}) and tools for knowledge mechanism analysis (\S \ref{appendix:Tools}).
Our contributions are as follows:


\input{fig/taxonomy}

\begin{itemize}[leftmargin=*,nosep]
    \item To the best of our knowledge, we are the first to review knowledge mechanisms in LLMs and provide \textbf{a novel taxonomy} across the entire life.
    \item We propose a new perspective to analyze knowledge utilization mechanisms from \textbf{three levels: memorization, comprehension and application, and creation}.
    \item We discuss \textbf{knowledge evolution} in individual and group LLMs, and analyze the inherent conflicts and integration in this process.
    \item We observe that LLMs have learned basic world knowledge. However, the learned knowledge is fragile, leading to challenges such as \textbf{hallucinations and knowledge conflicts}. We speculate that this fragility may be primarily due to improper learning data. Besides, the unlearned \textbf{dark knowledge} will exist long.
\end{itemize}

\paragraph{Comparison with Existing Surveys}
Previous interpretability surveys typically aim to investigate various \textit{methods for explaining the roles of different components} within LLMs from the global and local taxonomy \cite{DBLP:journals/corr/abs-2405-00208,DBLP:journals/tist/ZhaoCYLDCWYD24,DBLP:journals/corr/abs-2401-12874,DBLP:journals/corr/abs-1901-04592,PracticalReview,Review,PositionPaper,DBLP:journals/corr/abs-2402-01761,DBLP:journals/corr/abs-2407-02646}.
In contrast, this paper focuses on knowledge in LLMs.
Hence, \textit{our taxonomy, oriented from target knowledge in LLMs, reviews how knowledge is acquired, stored, utilized, and subsequently evolves}.
Additionally, previous taxonomy mostly explore the explainability \textit{during the inference stage} (a specific period), while ignoring knowledge acquisition during the pre-training stage and evolution during the post-training stage \cite{DBLP:conf/satml/RaukerHCH23,DBLP:journals/csur/LuoIHP24,DBLP:journals/coling/Apidianaki23,DBLP:journals/staeors/JiaoHLYMZYHYLMLCFTGZQWLBLSF23,DBLP:conf/satml/RaukerHCH23,rai2024practical}.
Our taxonomy aims to explore the \textit{dynamic evolution across all periods} from naivety to sophistication in both individual and group LLMs.
In contrast to the most similar survey \cite{DBLP:journals/ijautcomp/CaoLHS24} that introduces knowledge life cycle, our work focuses on the underlying mechanisms at each stage.

Generally, this paper may help us to explore and manipulate advanced knowledge in LLMs, examine current limitations through the history of knowledge evolution, and \textbf{inspire more efficient and trustworthy architecture and learning strategy for future models from knowledge mechanism perspective}.
Note that most hypotheses in this paper are derived from transformer-based LLMs.  
We also validate the generalizability of these hypotheses across other architectural models and then propose universality intelligence in \S \ref{appendix:Universality Intelligence}.

%% file: fig/taxonomy.tex
\tikzstyle{my-box}=[
    rectangle,
    draw=hidden-draw,
    rounded corners,
    text opacity=1,
    minimum height=1.5em,
    minimum width=5em,
    inner sep=2pt,
    align=center,
    fill opacity=.5,
]

\tikzstyle{utilization_leaf}=[my-box, minimum height=1.5em,
    fill=bluelight2!100, text=black, align=left,font=\scriptsize,
    inner xsep=2pt,
    inner ysep=4pt,
]
\tikzstyle{evolution_leaf}=[my-box, minimum height=1.5em,
    fill=greenlight, text=black, align=left,font=\scriptsize,
    inner xsep=2pt,
    inner ysep=4pt,
]

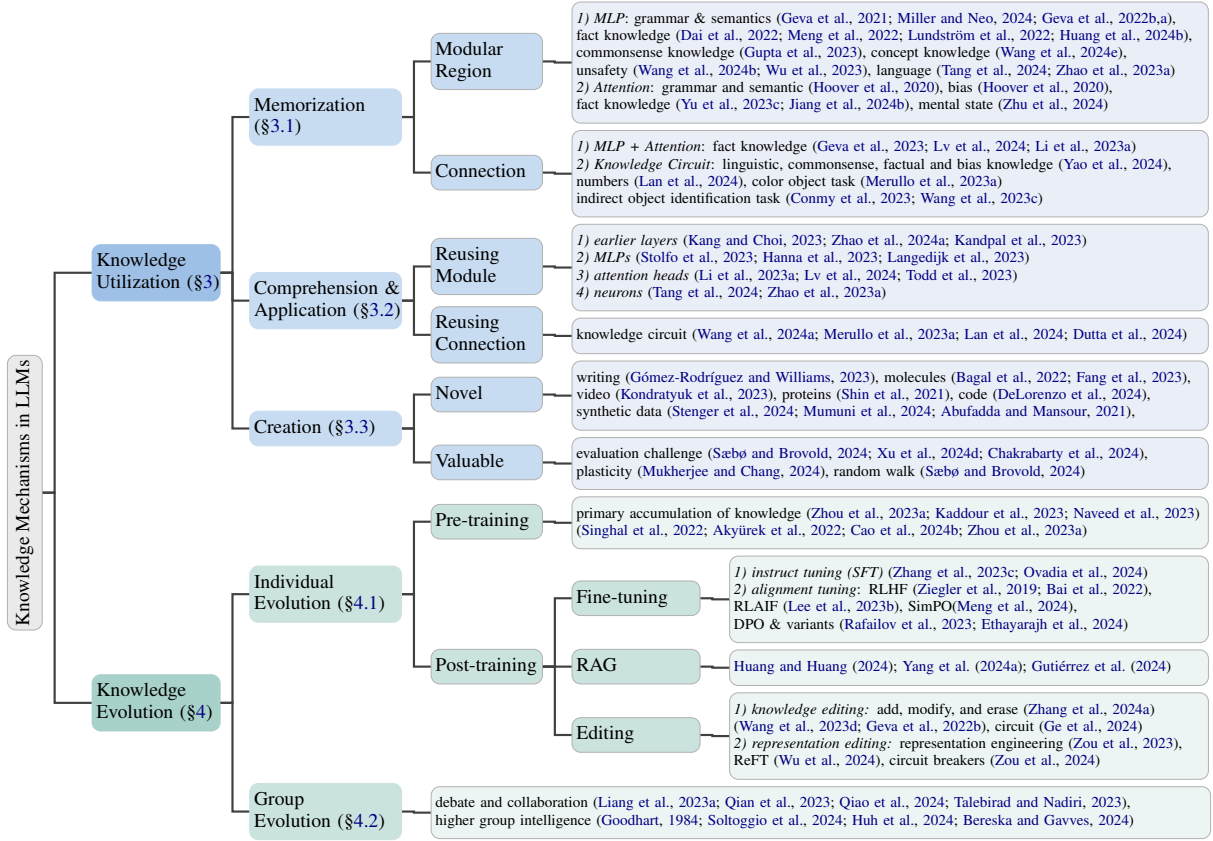
\begin{figure*}[!t]
    \centering
    \resizebox{\textwidth}{!}{
        \begin{forest}
            forked edges,
            for tree={
                grow=east,
                reversed=true,
                anchor=base west,
                parent anchor=east,
                child anchor=west,
                base=left,
                font=\small,
                rectangle,
                draw=hidden-draw,
                rounded corners,
                align=left,
                minimum width=4em,
                edge+={darkgray, line width=1pt},
                s sep=3pt,
                inner xsep=2pt,
                inner ysep=3pt,
                ver/.style={rotate=90, child anchor=north, parent anchor=south, anchor=center},
            },
            [
                Knowledge Mechanisms in LLMs, ver, color=hidden-draw, fill=mygray!30,
                text=black
                [
                    Knowledge \\ Utilization (\S \ref{Knowledge in LLMs}), color=myblue!100, fill=myblue!100, text width=5.0em, text=black
                    [
                        Memorization \\ (\S\ref{Memorization}), color=bluelight1!80, fill=bluelight1!80,  text width=6em, text=black
                            [
                                Modular \\Region, color=hidden-draw, fill=bluelight1!80,  text width=4.3em, text=black
                                [
                                {\textit{1) MLP}: grammar \& semantics \cite{Key-ValueMemories,An,DBLP:conf/emnlp/GevaCWG22,LM-Debugger}, 
                                \\fact knowledge \cite{KN,ROME,DBLP:conf/icml/LundstromHR22,DBLP:journals/corr/abs-2402-17700}, 
                                \\commonsense knowledge \cite{DBLP:conf/emnlp/GuptaMS00WT23}, concept knowledge \cite{ConceptEdit}, 
                                \\unsafety \cite{DINM,DBLP:conf/emnlp/WuLXDW0X23}, language \cite{Language-Specific,functionregion}
                                \\ \textit{2) Attention}: grammar and semantic \cite{DBLP:conf/acl/HooverSG20}, bias \cite{DBLP:conf/acl/HooverSG20},
                                \\fact knowledge \cite{Characterizing,dreamofelephants}, 
                                 mental state \cite{Beliefs}
                                } , utilization_leaf, text width=26.3em
                                ]
                            ]
                            [
                                Connection, color=hidden-draw, fill=bluelight1!80,  text width=4.3em, text=black
                                [
                                {
                                \textit{1) MLP + Attention}: fact knowledge \cite{geva-etal-2023-dissecting,DBLP:journals/corr/abs-2403-19521,DBLP:conf/nips/0002PVPW23}\\
                                \textit{2) Knowledge Circuit}: linguistic, commonsense, factual and bias knowledge \cite{knowledgecircuit}, \\numbers \cite{SharedCircuits}, color object task \cite{DBLP:journals/corr/abs-2310-08744}
                                \\indirect object identification task \cite{ACDC,WangVCSS23}
                                }, utilization_leaf, text width=26.3em
                                ]
                            ]
                    ]
                    [
                        Comprehension \& \\Application (\S \ref{Comprehension and Application}), color=bluelight1!80, fill=bluelight1!80, text width=6em, text=black
                        [
                            Reusing \\Module , color=hidden-draw, fill=bluelight1!80,  text width=4.3em, text=black
                                [
                                {
                                \textit{1) earlier layers} \cite{DBLP:conf/emnlp/KangC23,DBLP:journals/tist/ZhaoCYLDCWYD24,Long-TailKnowledge}
                                \\\textit{2) MLPs} \cite{DBLP:conf/emnlp/StolfoBS23,mathematical,DecoderLens}
                                \\\textit{3) attention heads} \cite{DBLP:conf/nips/0002PVPW23,DBLP:journals/corr/abs-2403-19521,DBLP:journals/corr/abs-2310-15213}
                                \\\textit{4) neurons} \cite{Language-Specific,functionregion}
                                }, utilization_leaf, text width=26.3em
                                ]
                        ]
                        [
                              Reusing \\Connection, color=hidden-draw, fill=bluelight1!80,  text width=4.3em, text=black
                            [
                            {
                            knowledge circuit \cite{Grokked,DBLP:journals/corr/abs-2310-08744,SharedCircuits,DBLP:journals/corr/abs-2402-18312}
                            }, utilization_leaf, text width=26.3em
                            ]
                        ] 
                    ]  
                    [
                        Creation (\S \ref{Creation}), color=bluelight1!80, fill=bluelight1!80, text width=6em, text=black
                        [
                            Novel, color=hidden-draw, fill=bluelight1!80,  text width=4.3em, text=black
                            [
                            {
                            writing \cite{DBLP:conf/emnlp/Gomez-Rodriguez23a}, molecules \cite{DBLP:journals/jcisd/BagalAVP22,DBLP:journals/corr/abs-2301-11259},
                            \\video \cite{DBLP:journals/corr/abs-2312-14125}, proteins \cite{Shin_Riesselman_Kollasch_McMahon_Simon_Sander_Manglik_Kruse_Marks_2021}, code \cite{DBLP:journals/corr/abs-2404-08806},
                            \\synthetic data \cite{DBLP:journals/jbd/StengerLFKB24,DBLP:journals/corr/abs-2403-10075,DBLP:conf/acit3/AbufaddaM21}, 
                            }, utilization_leaf, text width=26.3em
                            ]
                        ]
                        [
                            Valuable, color=hidden-draw, fill=bluelight1!80,  text width=4.3em, text=black
                            [
                            {evaluation challenge \cite{DBLP:journals/corr/abs-2403-06996,DBLP:journals/corr/abs-2401-11817,DBLP:conf/chi/ChakrabartyLAMW24},
                            \\plasticity \cite{DBLP:journals/corr/abs-2404-04436}, random walk \cite{DBLP:journals/corr/abs-2403-06996}
                            }, utilization_leaf, text width=26.3em
                            ]
                        ] 
                    ]  
                ]
                [
                   Knowledge \\Evolution (\S \ref{Knowledge Evolution}), color=mygreen!100, fill=mygreen!100, text width=5.0em, text=black
                       [
                            Individual \\Evolution (\S \ref{Individual Evolution}), color=mygreen!60, fill=mygreen!60, text width=6em, text=black
                            [
                                Pre-training, color=hidden-draw, fill=mygreen!60, text width=4.3em, text=black
                                [
                                {
                                primary accumulation of knowledge \cite{DBLP:conf/nips/ZhouLX0SMMEYYZG23, DBLP:journals/corr/abs-2307-10169, DBLP:journals/corr/abs-2307-06435}
                                \\\cite{DBLP:journals/corr/abs-2212-13138,akyurek2022towards,cao-etal-2024-retentive-forgetful,DBLP:conf/nips/ZhouLX0SMMEYYZG23}
                                }
                                    , evolution_leaf, text width=26.3em
                                ]
                           ] 
                           [
                                Post-training , color=hidden-draw, fill=mygreen!60, text width=4.3em, text=black
                                [
                                Fine-tuning, color=hidden-draw, fill=mygreen!60, text width=5em, text=black
                                    [
                                    {
                                    \textit{1) instruct tuning (SFT)} \cite{DBLP:journals/corr/abs-2308-10792,ovadia2024finetuning}
                                    \\\textit{2) alignment tuning}: RLHF \cite{DBLP:journals/corr/abs-1909-08593,DBLP:journals/corr/abs-2204-05862}, 
                                    \\RLAIF \cite{RLAIF}, SimPO\cite{DBLP:journals/corr/abs-2405-14734},
                                    \\DPO \& variants \cite{DBLP:conf/nips/RafailovSMMEF23,DBLP:journals/corr/abs-2402-01306}
                                    }, evolution_leaf, text width=19.7em
                                    ]
                                ]
                                [
                                RAG, color=hidden-draw, fill=mygreen!60, text width=5em, text=black
                                    [
                                    {
                                    \citet{DBLP:journals/corr/abs-2404-10981,DBLP:journals/corr/abs-2405-13021,DBLP:journals/corr/abs-2405-14831}
                                    }
                                        , evolution_leaf, text width=19.7em
                                    ]
                                ]
                                [
                                Editing, color=hidden-draw, fill=mygreen!60, text width=5em, text=black
                                    [
                                    {
                                    \textit{1) knowledge editing:} add, modify, and erase
                                    \cite{DBLP:journals/corr/abs-2401-01286}
                                    \\\cite{DBLP:journals/corr/abs-2310-16218,DBLP:conf/emnlp/GevaCWG22}, circuit \cite{ge2024circuits}
                                    \\\textit{2) representation editing:} representation engineering \cite{RepresentationEngineering}, 
                                    \\ReFT \cite{ReFT}, circuit breakers \cite{CircuitBreakers}
                                    }, evolution_leaf, text width=19.7em
                                    ]
                                ]
                           ] 
                       ]
                       [
                            Group \\Evolution (\S \ref{Group Evolution}), color=mygreen!60, fill=mygreen!60, text width=6em, text=black
                            [
                                {
                                debate and collaboration \cite{debate,chatdev,autoact,multi-agent-coll},
                                \\higher group intelligence \cite{goodhart,soltoggio2024collective,huh2024platonic,Review}
                                }
                                    , evolution_leaf, text width=32.3em
                                ]
                       ]
                ]
            ]   
        \end{forest}
    }
\caption{The taxonomy of knowledge mechanisms in LLMs.}
\label{taxonomy}
\end{figure*}

%% file: section/Preliminary.tex
\section{Preliminary}
\label{Preliminary}

\subsection{Knowledge Scope}
\label{Definition of Knowledge}

Knowledge is an awareness of facts, a form of familiarity, awareness, understanding, or acquaintance \cite{zagzebski2017knowledge, hyman1999knowledge,DBLP:journals/corr/abs-2301-06627,DBLP:conf/chi/GraySBM24}. 
It often involves the possession of information learned through experience and can be understood as a cognitive success or an epistemic contact with reality.
We denote a diverse array of knowledge as set $\mathbf{K}$, wherein each element $k \in \mathbf{K}$ is a specific piece of knowledge, which can be expressed by various records, e.g., a text record ``\textit{The president of the United States in 2024 is Biden}'' (denoted as $r_k$).

\subsection{Definition of Knowledge in LLMs}
\label{Definition of Knowledge within LLMs}
Given a LLM denoted as $\mathcal{F}$, we formulate that $\mathcal{F}$ master knowledge $k$ if $\mathcal{F}$ can correctly answer the corresponding question $r_{k\backslash t}$:
\begin{equation}
\begin{aligned}
&t = \mathcal{F}\left(r_{k\backslash t} \right) \\
&(t \in \mathbf{T}) \Rightarrow (\mathcal{F} \  \text{masters knowledge} \ k),
\end{aligned}
\label{knowledge}
\end{equation}
$t$ is the output of a LLM $\mathcal{F}$,
$r_{k\backslash t}$ is a record about knowledge $k$ that lacks pivot information.
Take an example for illustration: $r_{k\backslash t}$ is \textit{``The president of the United States in 2024 is \_\_''}, the pivot information is ``Biden''.
Note that, $r_{k\backslash t}$  can be represented by the above textual statement, captured through a question-answering pair (``\textit{Who is the President of the United States in 2024?}'', or conveyed by audio, video, image \footnote{While audio, video, and image records have been somewhat investigated, they are still relatively unexplored areas and thus are only discussed in \S \ref{Embodied Intelligence} and \S \ref{appendix:Knowledge Mechanisms in other Architectures}.}, and other equivalent expressions. 
The pivot information for $r_{k\backslash t}$ can be expressed by various fomarts, which are formulated as $\mathbf{T}= \left\{\text{``Biden''},\text{``Joe Biden''},\cdots\right\}$. 
If the output $t$ is an element from the correct answer set $\mathbf{T}$, we hypothesize that $\mathcal{F}$ master knowledge $k$.

\subsection{The Architecture of LLMs}
\label{The Structure of LLMs}
An LLM $\mathcal{F}$ consists of numerous neurons, which work systematically under a specific architecture.

\paragraph{Transformer-based architecture.} 
The prevailing architecture in current LLMs is the Transformer \cite{DBLP:conf/nips/VaswaniSPUJGKP17}. 
Specifically, a transformer-based LLM $\mathcal{F}$ begins with a token embedding, followed by $L$ layers transformer block, and ends with token unembedding used for predicting answer tokens.
Each transformer block layer $l$ consists of Attention Heads (Attention) and Multilayer Perceptron (MLP):
\begin{equation}
\begin{aligned}
  h_{l+1}=h_l+\text{MLP}\left( h_l+\text{Attention}\left( h_l \right) \right) , 
\end{aligned}
\label{transformer}
\end{equation}
$h_l$ is the hidden state from $l$-th layer.

\paragraph{Other architectures.} Other architectures including competitive variants of the transformer, e.g., SSM \cite{DBLP:journals/corr/abs-2312-00752}, TTT \cite{TTT} and RWKV \cite{DBLP:conf/emnlp/PengAAAABCCCDDG23}, and architectures in computer vision \cite{DBLP:journals/corr/abs-2308-09388} and multi-modal fields are detailed in \S \ref{appendix:Model Architecture}.

\subsection{Knowledge Analysis Methods}
\label{Methods of Knowledge Analysis}
Knowledge analysis method $\mathcal{M}$ aims to interpret how LLMs work inside and reveal precise causal connections between specific components and outputs \cite{Review}. 
Furthermore, if components $\mathbf{C}$ of $\mathcal{F}$ accurately infer $t$ through analysis method $\mathcal{M}$, it is assumed that the knowledge $k$ is presented by $\mathbf{C}$:
\begin{equation}
\begin{aligned}
&t = \mathcal{M}_{\mathbf{C}\subseteq \mathcal{F}}\left( r_{k\backslash t},\mathbf{C} \right),\\
&(t \in \mathbf{T}) \Rightarrow (\mathbf{C} \ \text{represents knowledge} \ k),
\end{aligned}
\label{store}
\end{equation}
The elements in set $\mathbf{C}$ may be individual neurons, MLPs, attention heads, a transformer block layer, or knowledge circuit \cite{knowledgecircuit}. 
These methods are divided into two categories: observation and intervention \cite{Review}.
\paragraph{Observation-based methods.} These methods aim to observe the internal information of $\mathcal{F}$, directly projecting the output of component $\mathbf{C}$ into human-understandable forms by $E$:
\begin{equation}
\begin{aligned}
&t = E_{\mathbf{C}\subseteq \mathcal{F}}\left( r_{k\backslash t},\mathbf{C}, \mathcal{F} \right), \\
\end{aligned}
\label{Observation-eq}
\end{equation}
$E$ is a evaluation metric, which can be a probe \cite{DBLP:conf/satml/RaukerHCH23}, logit lens \cite{logitlens}, or a sparse representation \cite{Scalingandevaluating}.
\textbf{Probe} is a meticulously trained classifier, and its classification performance is used to observe the relationship between model's behavior and the output of $\mathbf{C}$ \cite{DBLP:journals/coling/Belinkov22,DBLP:journals/tacl/ElazarRJG21,ChessKnowledge,DBLP:journals/corr/abs-2305-01610}.
\textbf{Logit lens} usually translate output of $\mathbf{C}$ into vocabulary tokens via token unembedding \cite{DBLP:conf/emnlp/GevaCWG22,DBLP:journals/corr/abs-2303-08112,DBLP:conf/conll/PalSYWB23,DBLP:conf/coling/DinKCG24,DecoderLens}.
\textbf{Sparse representation} maps the output of $\mathbf{C}$ into a higher-dimensional space with strong sparsity through dictionary learning \cite{DBLP:journals/corr/abs-2402-12201,olshausen1997sparse,DBLP:conf/acl-deelio/YunCOL21,karvonen2024measuring}, with sparse auto-encoder \cite{sparseautoencoders,DBLPjournals/corr/abs-2309-08600,DBLP:conf/nips/LeeBRN06,DBLP:journals/corr/abs-2406-04093} being a prominent example.
The higher-dimensional space represents independent (or monosemantic \cite{Monosemanticity}) and interpretable features more easily \cite{rai2024practical}.
The output of $\mathbf{C}$ is the combination \cite{ToyModels,Monosemanticity} of these features.


\paragraph{Intervention-based methods.} These methods allow for direct corruptions in LLMs to identify the critical $\mathbf{C}$ via intervention strategies $\mathcal{I}$.
Note that $\mathbf{C}$, encompassing various neuron combinations, correlates with specific model behaviors:
\begin{equation}
\begin{aligned}
&\mathbf{C} = \mathcal{I} \left( r_{k\backslash t},\mathcal{F} \right),\\
&t = E\left( r_{k\backslash t},\mathbf{C}, \mathcal{F} \right)\\
\end{aligned}
\label{Intervention-eq}
\end{equation}
$\mathcal{I}$ is also known as causal mediation analysis \cite{DBLP:conf/nips/VigGBQNSS20}, causal tracing \cite{ROME}, interchange interventions \cite{DBLP:conf/icml/GeigerWLRKIGP22}, activation patching \cite{WangVCSS23,DBLP:journals/corr/abs-2309-16042}, path patching \cite{DBLP:journals/corr/abs-2304-05969}, and causal scrubbing techniques \cite{CausalScrubbing}. 
Specifically, $\mathcal{I}$ consists of the following three steps.
1) \textbf{Clean run}:
$\mathcal{F}$ generates the correct answer $t$ based on the input $r_{k\backslash t}$.
2) \textbf{Corrupted run}: corrupt the generation process of $\mathcal{F}$ in the \textit{clean run} by introducing noise into the input or neurons \cite{ROME,DBLP:journals/corr/abs-2304-05969,DBLP:conf/emnlp/StolfoBS23,knowledgecircuit,ACDC,TransformerDebugger,DBLP:journals/corr/abs-2309-00244,DBLP:conf/blackboxnlp/HuangGDWP23}.
3) \textbf{Restoration run}: recover the correct answer $t$ by restoring unnoised information from $\mathbf{C}$ \cite{ROME,DBLP:conf/nips/VigGBQNSS20,WangVCSS23,DBLP:conf/iclr/ZhangBHRV17,Patching}.
For intervention-based methods, $E$ typically refers to the token unembedding used for predicting answer tokens.
Under the evaluation metric $E$, there exists a causal relationship between $\mathbf{C}$ and specific behavior of LLMs $\mathcal{F}$ in Eq \ref{Intervention-eq}.

%% file: section/Knowledge_Representation.tex
\section{Knowledge Utilization in LLMs}
\label{Knowledge in LLMs}

\begin{figure}[!t]
    \centering
    \includegraphics[width=0.5\textwidth]{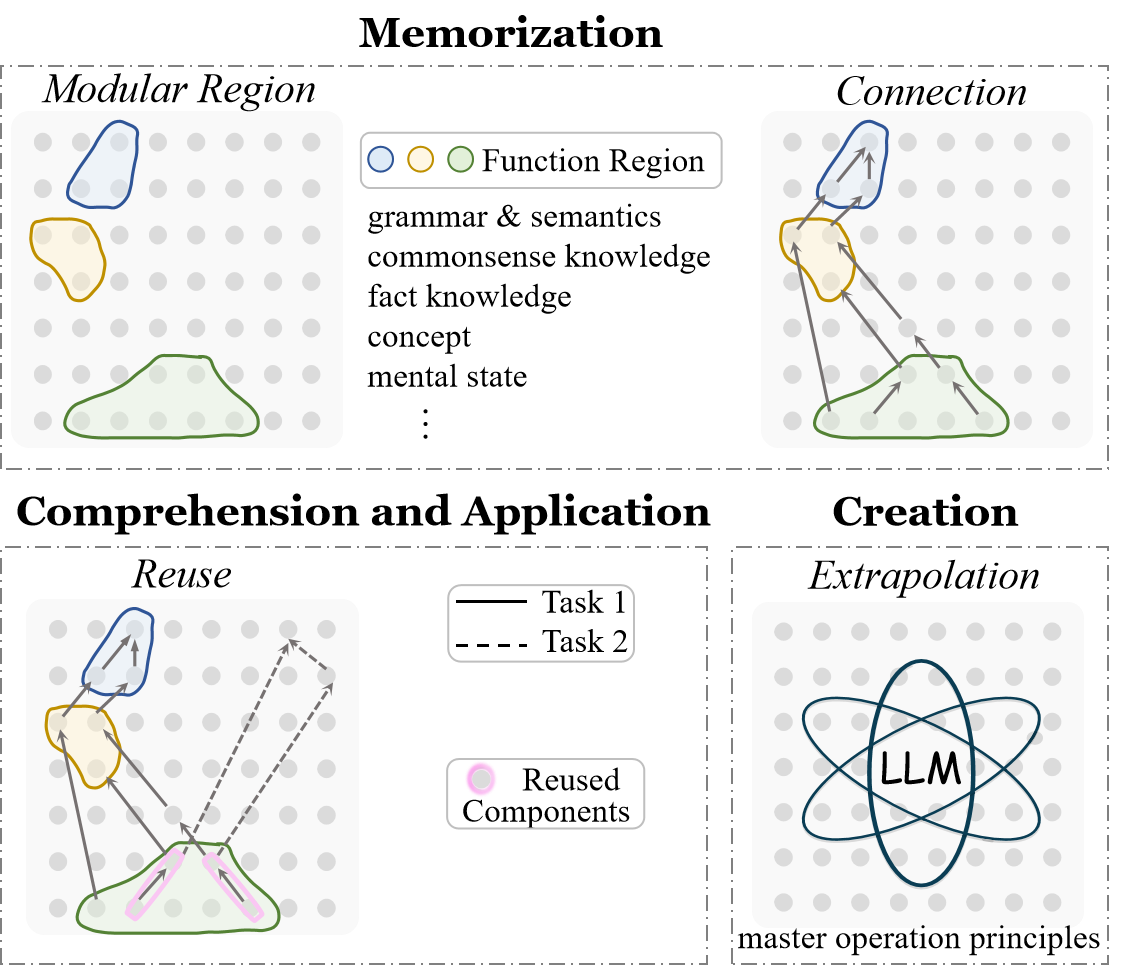}
    \caption{The mechanism analysis for knowledge utilization across three levels: memorization, comprehension and application, and creation.}
    \label{fig:utilization}
    \vskip -0.15in
\end{figure}

Inspired by Bloom's Taxonomy of cognition levels \cite{wilson2016anderson,bloom1956taxonomy,keene2010mapping,fadul2009collective}, we categorize knowledge representation and utilization within LLMs into three levels (as shown in Fig \ref{fig:utilization}): memorization, comprehension and application, and creation \footnote{
Note that we combine analyzing, evaluating, and creating from Bloom's Taxonomy into one category level (creation) in our taxonomy, as they are difficult to disentangle.
Specifically, creation emphasizes the capacity and process of forming \textit{novel} and \textit{valuable} things. 
Analyzing \cite{wilson2016anderson}, which breaks materials or concepts into parts, is used for creating \textit{novel} things.
Evaluating \cite{wilson2016anderson} is usually used for assessing the \textit{value} of new creations.}.
Note that these mechanistic analyses are implemented via methods in \S \ref{Methods of Knowledge Analysis}.
We further evaluate the applicability, advantages, and limitations of different methods in \S \ref{Comparison of Different Analysis Methods}.

\subsection{Memorization}
\label{Memorization}
Knowledge memorization \cite{DBLP:journals/corr/abs-2404-15146,MultifacetedMemorization} aims to remember and recall knowledge in the training corpus, e.g.,  specific terms (entities), grammar, facts, commonsense, concepts, etc \cite{DBLP:journals/corr/abs-2305-13673,DBLP:journals/corr/abs-2306-09296,DBLP:journals/corr/abs-2301-06627,DBLP:journals/corr/abs-2309-14316,DBLP:journals/corr/abs-2309-14402,DBLP:journals/corr/abs-2404-05405,DBLP:journals/ijautcomp/CaoLHS24}.
\textit{We posit knowledge memorization from Modular Region and Connection Hypothesis by reviewing existing research.}

\begin{bluebox}[Hypothesis 1: Modular Region]
Knowledge is Encoded in Modular Regions.
\label{Hypothesis 1}
\end{bluebox}

This modular region hypothesis simplifies knowledge representation in transformer-based models into isolated modular region, e.g., MLPs or {attention heads}.
\textbf{Knowledge is encoded via MLPs.} \citet{Key-ValueMemories} posit that MLPs operate as key-value memories and each individual key vector corresponds to a specific \textit{semantic pattern} or \textit{grammar}.
Based on the above finding, \citet{DBLP:conf/emnlp/GevaCWG22,LM-Debugger} reverse engineer the operation of the MLPs layers and find that MLPs can promote both {semantic} (e.g., measurement semantic including kg, percent, spread, total, yards, pounds, and hours) and \textit{syntactic} (e.g., adverbs syntactic including largely, rapidly, effectively, previously, and normally) {concepts} in the vocabulary space.
\citet{An} find a single MLP neuron (in GPT-2 Large) capable of generating ``an'' or ``a''.
Subsequently, \textit{fact} \cite{KN,ROME} and \textit{commonsense} knowledge \cite{DBLP:conf/emnlp/GuptaMS00WT23} are found. 
Advanced language-specific neurons \cite{Language-Specific}, linguistic regions \cite{functionregion}, entropy neurons \cite{stolfo2024confidence},  query-relevant neurons \cite{Query-Relevant}, abstract conceptual \cite{ConceptEdit} and unsafe \cite{DINM,DBLP:conf/emnlp/WuLXDW0X23} knowledge, are also observed in MLPs.
In addition to MLP, \textbf{knowledge is also conveyed by attention heads} \cite{geva-etal-2023-dissecting,SuccessorHeads}.
\citet{DBLP:conf/acl/HooverSG20} explain the knowledge each attention head has learned. Specifically, attention heads store evident \textit{linguistic features, positional information}, and so on.
Besides, \textit{fact knowledge} \cite{Characterizing,DBLP:conf/nips/0002PVPW23} and \textit{bias} \cite{DBLP:conf/acl/HooverSG20} are mainly convey by attention heads.
\citet{dreamofelephants} further observe that LLMs leverage self-attention to gather information through certain tokens in the contexts, which serve as clues, and use the value matrix for associative memory.
Later, \citet{Beliefs} also find that attention heads can simulate \textit{mental stat}e and activate ``Theory of Mind'' (ToM) capability.

However, Hypothesis 1 ignores the connections between different regions.
Inspired by advancements in neuroscience \cite{thiebaut2022emergent}, Hypothesis 2 asserts that the connection of different components integrates knowledge, rather than the isolated regions in Hypothesis 1.


\begin{bluebox}[Hypothesis 2: Connection]
Knowledge is Represented by Connections.
\end{bluebox}

\citet{geva-etal-2023-dissecting} outline the encoding of factual knowledge (e.g., ``The capital of Ireland is Dublin'') through the following three steps: (1) subject (Ireland) information enrichment in MLPs, (2) the relation (capital of) propagates to the last token, (3) object (Dublin) is extracted by attention heads in later layers.
This claim is supported by \citet{PMET}.
Similarly, \citet{DBLP:journals/corr/abs-2403-19521} conclude that task-specific attention head may move the topic entity to the final position of the residual stream, while MLPs conduct relation function.
Moreover, the recent prominent knowledge circuit framework \cite{DBLP:journals/corr/abs-2401-03646,knowledgecircuit,DBLP:journals/corr/abs-2402-12201,elhage2021mathematical,DBLP:journals/corr/abs-2403-19647} advocates leveraging a critical computational subgraph among all components to explore internal knowledge within LLM parameters.
The competencies for indirect object identification and color object tasks are discovered to be embedded in specialized knowledge circuits \cite{ACDC,WangVCSS23,DBLP:journals/corr/abs-2310-08744,yu2024functional}.
\citet{SharedCircuits} also identify number-related circuits that encode the predictive ability of Arabic numerals, number words, and months.
More importantly, experimental evidence demonstrates that various types of knowledge, including linguistic, commonsense, factual, and biased information, are encapsulated in specific knowledge circuits \cite{knowledgecircuit}.
Interestingly, knowledge encoded by specific circuits can rival or even surpass that of the entire LLM.
This may be because knowledge circuits memorized the relevant knowledge, while noise from other components might impede the model's performance on these tasks.


\subsection{Comprehension and Application}
\label{Comprehension and Application}
Knowledge comprehension and application focus on demonstrating the understanding of memorized knowledge and then solving problems in new situations, e.g., \textit{generalization on out-of-domain tasks} \cite{Grokked}, \textit{reasoning \cite{DBLP:conf/emnlp/HouLFSZZBS23} and planning} \cite{ChessKnowledge}.
\citet{DBLP:journals/corr/abs-2303-11873} denote the transition from memorization to comprehension and application as grokking, and suggest that the grokking derives from two largely distinct subnetworks competition.
Intuitively, only knowledge that is correctly memorized \cite{MultifacetedMemorization} in \S \ref{Memorization} can be further applied to solving complex tasks.
Therefore, we posit the following Reuse Hypothesis from two knowledge memorization perspectives.

\begin{bluebox}[Hypothesis 3: Reuse]
LLMs Reuse Certain Components during Knowledge Comprehension and Application.
\end{bluebox}

\textbf{From the Modular Region Perspective, knowledge utilization reuses some regions.}
These regions might include a few neurons, attention heads, MLPs, a transformer layer, or partial knowledge circuits.
Generally, basic knowledge (position information, n-gram pattern, syntactic features) tends to be stored at earlier layers, while sophisticated knowledge (mental state, emotion, and abstract concept, e.g., prime number, Camelidae, and safety) is located at later layers \cite{Beliefs,DBLP:journals/corr/abs-2404-07066,DINM,ConceptEdit,ShortGPT,kobayashi2023analyzing}. 
Therefore, \textit{neurons of earlier layers related to basic knowledge tend to be reused} \cite{DBLP:conf/emnlp/KangC23,DBLP:journals/tist/ZhaoCYLDCWYD24,Long-TailKnowledge}.
Various math reasoning tasks also utilize the attention mechanism in initial layers to map input information to the final token positions, subsequently generating answers using a set of MLPs in later layers \cite{DBLP:conf/emnlp/StolfoBS23,mathematical,DecoderLens}.
Besides, \textit{some specific function regions are also reused}.
Specifically, retrieval heads \cite{DBLP:conf/nips/0002PVPW23} are reused for Chain-of-Thought (CoT) reasoning and long-context tasks. These retrieval heads are found in 4 model families, 6 model scales, and 3 types of fine-tuning.
Subsequently, induction heads, identified in Llama and GPT, are claimed to be reused for in-context learning (ICL) tasks \citet{InductionHeads,InductionHeadsPattern}.
Attention heads can map country names to their capitals in capital city-related tasks \cite{DBLP:journals/corr/abs-2403-19521}. 
Language-specific neurons (in Llama and BLOOM) are responsible for multiple language related tasks, such as English, French, Mandarin, and others \citet{Language-Specific}.
\citet{functionregion} further reveal linguistic regions (in Llama) correspond to linguistic competence, which is the cornerstone for performing various tasks.
Later, function regions related to the process of math reasoning are also discovered in LLMs.
For instance, the last layer of GPT-2 (trained from scratch) has been observed to exhibit mathematical reasoning abilities across various math questions \cite{Part2.1}.
\textbf{From the Connection Perspective, knowledge utilization shares partial knowledge circuits.}
For instance, similar tasks share subgraphs (computational circuits) with analogous roles \cite{SharedCircuits}.
Besides, knowledge circuits (in GPT2) are reused to solve a seemingly different task, e.g., indirect object identification and colored objects tasks \cite{DBLP:journals/corr/abs-2310-08744}.
\citet{Grokked} further observe that two-hop composition reasoning tasks reuse the knowledge circuits from the first hop. 
\citet{knowledgecircuit} also believe that this reuse phenomenon exists in factual recall and multi-hop reasoning. Specifically, sub-circuits are reused in similar factual knowledge, such as tasks related to ``city\_in\_country'', ``name\_birth\_place'', and ``country\_language''.
Besides, \citet{DBLP:journals/corr/abs-2402-18312} demystify LLMs how to perform CoT reasoning, i.e., Llama facilitates CoT tasks via multiple parallel circuits enjoying significant intersection.

\subsection{Creation}
\label{Creation}
Knowledge creation \cite{runco2012standard,sternberg2006nature} emphasizes the capacity and process of forming \textit{novel} and \textit{valuable} things, rather than the existing ones (i.e., LLMs have seen) discussed in \S \ref{Memorization} and \S \ref{Comprehension and Application}.
The creations encompass two levels:
1) LLMs create new terms following the current world's principles comprehended by LLMs, such as new proteins \cite{Shin_Riesselman_Kollasch_McMahon_Simon_Sander_Manglik_Kruse_Marks_2021}, molecules \cite{DBLP:journals/jcisd/BagalAVP22,DBLP:journals/corr/abs-2301-11259,DBLP:conf/emnlp/EdwardsLRHCJ22}, code \cite{DBLP:journals/corr/abs-2404-08806}, video \cite{DBLP:journals/corr/abs-2312-14125}, models \cite{DBLP:journals/corr/abs-2404-16792}, names for people and companies, written stories \cite{DBLP:journals/corr/abs-2405-13012,DBLP:conf/emnlp/Gomez-Rodriguez23a,DBLP:journals/corr/abs-2405-01660}, synthetic data \cite{DBLP:journals/jbd/StengerLFKB24,DBLP:journals/corr/abs-2403-10075,DBLP:conf/acit3/AbufaddaM21}, etc.
These novel items operate according to the existing rules, e.g., law of conservation of energy, reasoning logic \cite{Grokked}, or principles of probability theory.
2) LLMs may generate new rules, such as mathematical theorems, and the resulting terms will operate according to the new rules.
We posit that the knowledge creation of LLMs may derive from the Extrapolation Hypothesis.

\begin{bluebox}[Hypothesis 4: Extrapolation]
LLMs May Create Knowledge via Extrapolation.
\end{bluebox}

The expression of knowledge is diverse; some knowledge is inherently continuous. Therefore, it is difficult, if not impossible, to represent certain knowledge using discrete data points \cite{continuity,penroselimitations,markman2013knowledge}. 
LLMs utilize insights into the operational principles of the world to extrapolate additional knowledge from known discrete points, bridging gaps in knowledge and expanding our understanding of the world\cite{heilman2003creative,DBLP:journals/corr/abs-2401-17692,DBLP:journals/corr/abs-2311-03658,DBLP:journals/corr/abs-2312-14125}.
Drawing inspiration from research on human creativity \cite{DBLP:journals/corr/abs-2303-12003}, the \textit{physical implementation of knowledge extrapolation relies on the plasticity of neurons} \cite{DBLP:journals/corr/abs-2404-04436}. 
Specifically, plasticity refers to LLMs changing activations and connectivity between neurons according to the input \cite{Gaming}.

However, from statistical perspective, the intricate connections and activations between neurons, though not infinite, resist exhaustive enumeration. In terms of value, not all creations are valuable. 
Obtaining something valuable with an exceedingly low probability is impractical, as even a monkey could theoretically print Shakespeare's works.
How do LLMs ensure the probability of generating valuable creations? 
\textit{What are the mechanisms underlying the novelty and value of creation?}
A prevalent conjecture posits that \textbf{novelty is generated through the random walk} \cite{DBLP:journals/corr/abs-2403-06996}.
However, intuitively, \textbf{current LLMs themselves seem unable to evaluate the value of creations due to architectural limitations} \cite{DBLP:conf/chi/ChakrabartyLAMW24}. 
Because, once the next token is generated, there is no intrinsic mechanism for accepting or rejecting the creations. This hinders the evaluation of the usefulness and value of proposed novelties, as humans do, by bending, blending, or breaking biases \cite{DBLP:journals/corr/abs-2403-06996}.
Some works assume that each token is indeed valuable and meets long-term expectations.
However, the well-known hallucination problem \cite{DBLP:journals/corr/abs-2401-11817} of LLMs refutes this assumption.
Besides, the transformer architecture struggles with long context \cite{DBLP:journals/corr/abs-2404-02060}, despite the existence of many variants for addressing this issue \cite{DBLP:journals/corr/abs-2311-12351,InfiniteContext}.
More importantly, MLPs of Transformer may also work contrary to creativity, i.e., the increased attentions narrow the conditional distribution for token prediction \cite{DBLP:journals/corr/abs-2403-06996}.


\subsection{Comparison of Different Mechanism Analysis Methods}
\label{Comparison of Different Analysis Methods}
The above four Hypotheses are achieved by Observation-based and Intervention-based methods.
These two methods are typically combined to trace knowledge in LLMs \cite{TransformerDebugger,ghandeharioun2024patchscopes}.
Most knowledge analysis methods are architecture-agnostic and can be adapted to various models.

Each method is suitable for different scenarios.
Specifically, the Modular Region Hypothesis can be analyzed using either Observation-based or Intervention-based methods. In contrast, the Connection Hypothesis, which examines inter-regional connectivity, generally necessitates Intervention-based methods.
However, the results of knowledge mechanism analysis depend heavily on different methods and are sensitive to evaluation metrics and implementation details \cite{DBLP:conf/nips/SchwettmannSMCL23}.
Hence, \citet{DBLP:journals/corr/abs-2402-17700} propose a dataset, RAVEL, to quantify the comparisons between a variety of existing interpretability methods. They suggest that methods with supervision are better than methods with unsupervised featurizers.
Later, \citet{DBLP:journals/corr/abs-2309-16042} further systematically examine the impact
of methodological details in intervention-based methods.
For corrupted run, they recommend Symmetric Token Replacement (e.g., ``The Eiffel Tower''$\rightarrow$``The Colosseum'') \cite{DBLP:journals/corr/abs-2404-03646,DBLP:conf/nips/VigGBQNSS20} {instead of Gaussian Noising} \cite{ROME}, which disrupts the model's internal mechanisms.
For metric $E$, both logit lens and probe can be employed to trace factual knowledge \cite{ROME}, where the target output is typically few tokens. 
In this scenario, \citet{DBLP:journals/corr/abs-2309-16042} advocate using the logit lens over probes for evaluation metric $E$ due to its fine-grained control over localization outcomes.
Moreover, probe is capable of exploring abstract knowledge and abilities, such as theory of mind or mental states \cite{Beliefs,Part2.1,jin2024latent}, where the target output requires multiple tokens to express.
\citet{jin2024latent} suggest that deeper probes are more (generally) more accurate.

%% file: section/Evolution.tex
\section{Knowledge Evolution in LLMs}
\label{Knowledge Evolution}

Knowledge in LLMs should evolve with changes in the external environment. We introduce the Dynamic Intelligence Hypothesis for knowledge evolution in individuals and groups.

\begin{bluebox}[Hypothesis 5: Dynamic Intelligence]

Conflict and Integration Coexist in the Dynamic Knowledge Evolution of LLMs.
\end{bluebox}

\subsection{Individual Evolution}
\label{Individual Evolution}
Immersed in a dynamic world, individuals mature through an iterative process of memorization, forgetting, error correction, and deepening understanding of the world around them. 
Similarly, LLMs dynamically encapsulate knowledge into parameters through the process of conflict and integration.

In the \textit{pre-training phase}, LLMs start as blank slates, facilitating easier acquisition for new knowledge \cite{DBLP:journals/corr/abs-2404-05405}. Consequently, numerous experiments demonstrate that LLMs accumulate vast amounts of knowledge during this stage \cite{cao-etal-2024-retentive-forgetful,DBLP:conf/nips/ZhouLX0SMMEYYZG23, DBLP:journals/corr/abs-2307-10169, DBLP:journals/corr/abs-2307-06435,DBLP:journals/corr/abs-2212-13138}.
\citet{akyurek2022towards} delve further into identifying which training examples are instrumental in endowing LLMs with specific knowledge.
However, contradictions during the pre-training stage may induce conflicts among internal parametric knowledge.
On the one hand, the false and contradictory information in training corpus propagate and contaminate related memories in LLMs via semantic diffusion, introducing broader detrimental effects beyond direct impacts \cite{Spread}.
On the other hand, LLMs tend to prioritize memorizing more frequent and challenging facts, which can result in subsequent facts overwriting prior memorization, significantly hindering the memorization of low-frequency facts \cite{lu2024scaling}.
In other words, LLMs struggle with balancing and integrating both low and high-frequency knowledge.

After pre-training, LLMs are anticipated to refresh their internal knowledge to keep pace with the evolving world during \textit{post-training stage}. 
Although LLMs seem to absorb new knowledge through continued learning, follow user instructions via instruct tuning \cite{DBLP:journals/corr/abs-2308-10792}, and align with human values through alignment tuning \cite{DBLP:journals/corr/abs-1909-08593}, \citet{ji2024language} have noted that LLMs intrinsically resist alignment during the post-training phase.
In other words, LLMs tend to learn factual knowledge through pre-training, whereas fine-tuning \footnote{Fine-tuning includes instruct tuning and alignment tuning \cite{zhao2023survey}.} teaches them to utilize it more efficiently \cite{DBLP:journals/corr/abs-2405-05904,DBLP:conf/nips/ZhouLX0SMMEYYZG23,ovadia2024finetuning}.
\citet{ren2024learning} also posit that instruction tuning is a form of self-alignment with existing internal knowledge rather than a process of learning new information. 
We conjecture that the debate on whether these processes truly introduce new knowledge stems from information conflicts. For example, the conflict between outdated information within LLMs and new external knowledge exacerbates their difficulty in learning new information.
To mitigate information conflicts, \citet{DBLP:journals/corr/abs-2311-08011} propose first forgetting old knowledge then learning new knowledge.
Another technique, retrieval-augmented generation (RAG) \cite{DBLP:journals/corr/abs-2404-10981}, while avoiding conflicts within internal parameters, still needs to manage conflicts between retrieved external information and LLMs' internal knowledge \cite{DBLP:journals/corr/abs-2403-08319}.
RAG also attempt to efficiently and effectively integrate new knowledge across passages or documents using multiple retrieval \cite{DBLP:journals/corr/abs-2405-13021} and hippocampal indexing \cite{DBLP:journals/corr/abs-2405-14831}.
Besides, editing technologies, including knowledge and representation editing, exhibit promising potential for knowledge addition, modification, and erasure.
Specifically, knowledge editing \cite{ROME,DBLP:conf/iclr/MitchellLBFM22,DBLP:conf/emnlp/CaoAT21,DBLP:journals/corr/abs-2401-01286,DBLP:journals/corr/abs-2310-16218,DBLP:journals/corr/abs-2310-19704} aims to selectively modify model parameters responsible for specific knowledge retention, while representation editing \cite{RepresentationEngineering,ReFT} adjusts the model’s conceptualization of knowledge to revise the stored knowledge within LLMs.
Note that the other strategy for knowledge editing adds external parameters or memory banks for new knowledge while preserving models’ parameters. 
We also provide the comparison of the above methods in \S \ref{appendix:Comparison of Methods for Knowledge Evolution} for better understanding.

\subsection{Group Evolution}
\label{Group Evolution} 

Besides individual learning, social interaction plays a pivotal role in the acquisition of new knowledge and is a key driver of human societal development~\cite{baucalsocial,levine1993social}. 
LLMs, also known as agents, collaborate to accomplish complex tasks during group evolution, each bearing unique knowledge that may sometimes contradict each other.
Therefore, contrary to individual evolution, \textit{group evolution encounters intensified conflicts, such as conflicts in specialized expertise among agents, competing interests, cultural disparities, moral dilemmas, and others}.
To achieve consensus and resolve conflicts, agents must first clarify their own and others' goals (beliefs) through internal representations in models \cite{Beliefs,RepresentationEngineering}.
Agents then discuss, debate, and reflect on shared knowledge through various communication methods \cite{DBLP:conf/iclr/ChanCSYXZF024,DBLP:conf/icml/SmitGDBP24,DBLP:journals/corr/abs-2406-11776,soltoggio2024collective}, e.g., prompt instructions, task and agent descriptions, parameter signals (activation and gradient), and representations of models.
However, conformity of agents, which tends to believe the majority's incorrect answers rather than maintaining their own, hinders conflict resolution during group evolution \cite{jintian-acl,DBLP:conf/www/MaRCS24}.
Note that the group also struggles with automating moral decision-making when facing moral conflicts.
Specifically, agents in the group miss ground truth for moral ``correctness'' and encounter dilemmas due to changes in moral norms over time \cite{DBLP:journals/aiethics/HagendorffD23}.
Generally, when, what, and how to share knowledge in the communication process to maximize learning efficiency and long-term expectations are still open questions in group evolution.

Through debate and collaboration, \textit{groups integrate more knowledge and can surpass the cognition of individual units} \cite{debate,chatdev,autoact,multi-agent-coll,jintian-acl}.
This derives from the assumption that each individual unit can contribute to
and benefit from the collective knowledge \cite{soltoggio2024collective,DBLP:journals/corr/abs-2401-11839}.
In addition, \textit{``When a measure becomes a target, it ceases to be a good measure''}, which implies that optimizing one objective on a single individual will inevitably harm other optimization objectives to some extent. 
Hence, it is unrealistic for an individual to learn all knowledge compared to group optimization.
Interestingly, LLM groups also follow the collaborative scaling law \cite{Collaboration}, where normalized solution quality follows a logistic growth pattern as scaling agents.
Moreover, some works \cite{huh2024platonic,Review} propose that knowledge tends to converge into the same representation spaces among the whole artificial neural models group with different data, modalities, and objectives.

\subsection{Comparison of Different Evolution Strategies}
\label{Comparison of Different Evolution Strategies}
Individuals and groups achieve dynamic intelligence primarily through two strategies: updating internal parametric knowledge \cite{DBLP:conf/nips/ZhouLX0SMMEYYZG23,autoact} and leveraging external knowledge \footnote{
Leveraging external knowledge includes using prompts \cite{DBLP:journals/corr/abs-2402-04559}, ICL, and RAG.} \cite{DBLP:journals/corr/abs-2404-10981,DBLP:journals/corr/abs-2402-04559}.
These two strategies are usually used together in applications \cite{ExplicitMemory}.

\textit{Updating internal parametric knowledge} necessitates high-quality data for parameter adjustments \cite{vashishtha2024teaching,DBLP:journals/ijautcomp/CaoLHS24}. 
Data proves pivotal when fine-tuning models to acquire new knowledge.
\citet{ovadia2024finetuning} also posit that the continued training of LLMs via unsupervised tuning generally exhibits suboptimal performance when it comes to acquiring new knowledge. 
Note that updating internal parametric knowledge requires resolving conflicts among internal parameters.
The crux of effective internal knowledge updating lies in preserving the consistency of the model’s parameter knowledge before and after tuning.
In contrast, \textit{leveraging external knowledge} requires managing conflicts within the external knowledge itself \footnote{Inconsistencies in external information are common, as external documents often contain conflicting data, particularly in contexts for RAG.} as well as conflicts between external and internal knowledge \cite{DBLP:journals/corr/abs-2403-08319,DBLP:journals/corr/abs-2405-18727}.
Besides, parametric knowledge compresses extensive information, promoting grokking and enhancing generalization \cite{Grokked}. 
In contrast, leveraging external knowledge avoids high training costs but necessitates substantial maintenance and retrieval costs for every user query.
Therefore, the \textit{combination of these two strategies} is promising.
An attempt for combination \cite{ExplicitMemory} suggests employing RAG for low-frequency knowledge and parametric strategy for high-frequency knowledge.

%% file: section/Application.tex
\section{Application of Knowledge Mechanism}
\label{Application}
The mechanism analysis of knowledge utilization and evolution may provide an avenue to construct more efficient and trustworthy models in practice.

\subsection{Efficient LLMs}
\label{Efficient LLMs}

Researchers have been working to reduce the cost of training and inference for LLMs through various optimization strategies, including architecture \cite{DBLP:conf/emnlp/AinslieLJZLS23,DBLP:journals/jmlr/FedusZS22}, data quality \cite{DBLP:journals/corr/abs-2304-08442}, parallelization \cite{DBLP:journals/corr/abs-2401-10241}, generalization theory \cite{DBLP:journals/corr/abs-2405-05409}, hardware \cite{DBLP:journals/corr/abs-2304-03208}, scaling laws \cite{DBLP:journals/corr/abs-2203-15556}, optimizer \cite{DBLP:journals/corr/abs-2305-14342}, etc.
The underlying knowledge mechanisms offer LLMs new potential for efficiently storing, utilizing, and evolving knowledge.

For \textbf{knowledge storage and utilization} in LLMs, \textit{knowledge (memory) circuit} provides the theory to decompose the knowledge computations of an LLM into smaller, recurring parts \cite{ExplicitMemory}.
These smaller parts guide the determination of which types of knowledge should be encoded into parameters. 
Therefore, $\text{Memory}^3$ \cite{ExplicitMemory} designs an explicit memory mechanism for Transformer-based LLMs, alleviating the burden of parameter size.
Specifically, $\text{Memory}^3$ designs external information, explicit memory, and implicit memory for different usage frequencies, reducing writing and reading costs.
For \textbf{knowledge evolution}, the knowledge mechanism analysis inspires \textit{editing} and \textit{model merging}.
The details of editing technologies can be found in \S \ref{Group Evolution}.
Model merging technologies \footnote{Model merging also includes methods that directly interpolation \cite{DBLP:journals/corr/abs-2403-13257} or randomly fusing \cite{DBLP:journals/corr/abs-2311-03099}, ignoring parameter directions. These methods are naive and are not the focus of our discussion here.} leverage parameter directions to combine multiple task-specific models into a single multitask model without performing additional training rather than training from scratch.
For instance, 
Task Arithmetic \cite{DBLP:conf/iclr/IlharcoRWSHF23} identifies the weight directions of task capabilities in different models, and then integrates a more powerful model by arithmetic operations on weight directions.
TIES \cite{DBLP:conf/nips/YadavTCRB23} resolves parameters directions conflicts, and merges only the parameters that are in alignment with the final agreed-upon sign.
\citet{DBLP:journals/corr/abs-2403-13187} further propose evolutionary optimization of model merging, which automatically discovers effective combinations of open-source models, harnessing their group intelligence without requiring extensive training data or computational resources.
Besides, the Lottery Ticket Hypothesis \cite{DBLP:conf/iclr/FrankleC19} provides a cornerstone for \textit{model compression}, generalizing across various datasets, optimizers, and model architectures \cite{DBLP:conf/nips/MorcosYPT19,DBLP:conf/nips/ChenCWGLW21}.
However, model compression often limits the success of editing and model merging \cite{kolbeinsson2024composable}.
This phenomenon poses challenges for practical implementations, highlighting the need for more effective strategies.

\subsection{Trustworthy LLMs}
\label{Trustworthy LLMs}

Numerous studies investigate the underlying causes of security risks \cite{reuel2024open,ren2024safetywashing,li2024more,bengio2024government,bengio2024managing,DBLP:journals/corr/abs-2405-06624}.
In particular,
\citet{DBLP:conf/nips/0001HS23} delve into the safety of LLM and reveal that the success of jailbreak is mainly due to the distribution discrepancies between malicious attacks and training data.
\citet{DBLP:conf/emnlp/GevaCWG22} and \citet{DINM} further discover that some parameters within LLMs, called toxic regions, are intrinsically tied to the generation of toxic content. 
\citet{ji2024language} even conjecture that LLMs resist alignment.
Therefore, traditional aligned methods, DPO \cite{DBLP:conf/nips/RafailovSMMEF23} and SFT, seem to merely bypass toxic regions \cite{DBLP:journals/corr/abs-2401-01967,DINM}, making them susceptible to other jailbreak attacks \cite{DBLP:journals/corr/abs-2311-09096}.

Inspired by the knowledge mechanism analysis in LLMs, a promising trustworthy strategy may be \textbf{designing architecture and training process} during the pre-training phase to encouraging modularity \cite{DBLP:journals/entropy/LiuGT24,DBLP:journals/corr/abs-2310-07711}, sparsity \cite{DBLPconf/icml/ChughtaiCN23}, and monosemanticity \cite{Monosemanticity,DBLP:journals/corr/abs-2211-09169}, which make the reverse engineering process more tractable \cite{DBLP:journals/corr/abs-2211-09169,Monosemanticity,DBLP:journals/entropy/LiuGT24,DBLP:journals/corr/abs-2310-17230}.
Yet, maintaining sparsity for a vast amount of world knowledge requires substantial resources, and whether monosemantic architecture can support advanced intelligence remains elusive.
Besides, \textbf{machine unlearning} \cite{nguyen2022survey,tian2024forget,DBLP:journals/corr/abs-2310-10683} aims to forget privacy or toxic information learned by LLMs. 
However, these unlearning methods suffer overfitting, forgetting something valuable due to the difficulty of disentangling verbatim memorization and general capabilities \cite{VerbatimMemorization,DBLP:journals/corr/abs-2404-02062}.
Another alternative technique is \textbf{knowledge editing}, precisely modifying LLMs using few instances during the post-training stage \cite{DBLP:journals/corr/abs-2310-19704,yao-etal-2023-editing,DBLP:journals/corr/abs-2310-16218,Fundamental,DBLP:journals/corr/abs-2402-19465}.
Extensive experiments demonstrate that knowledge editing has the potential to detoxify LLMs \cite{DBLP:journals/corr/abs-2402-13462}.
Specifically, \cite{DBLP:conf/emnlp/WuLXDW0X23} and \citet{DBLP:conf/emnlp/GevaCWG22} deactivate the neurons related to privacy information and toxic tokens, respectively.
\cite{DINM} identify and then erases toxic regions in LLMs. 
However, knowledge editing also introduces side effects, such as the inability of the modified knowledge to generalize to multi-hop tasks \cite{DBLP:conf/emnlp/ZhongWMPC23,DBLP:journals/corr/abs-2310-02129,DBLP:journals/corr/abs-2307-12976,kong2024aligning} and the potential to impair the model's general capabilities \cite{DBLP:journals/corr/abs-2401-04700,HengJi}. 
Therefore, recent efforts focus on \textbf{representation editing} instead of editing parameters in knowledge editing \cite{RepresentationEngineering,DBLP:journals/corr/abs-2308-10248,AlignmentHiddenStates,Beliefs}.
These representations (hidden states) within LLMs can trace and address a wide range of safety-relevant problems, including honesty, harmlessness, and power seeking.
Later, \cite{ReFT} develop a family of representation finetuning methods to update new knowledge. \cite{CircuitBreakers} propose circuit-breaking \cite{DBLP:journals/corr/abs-2309-05973}, directly controlling the representations that are responsible
for harmful outputs.
However, these representation editing strategies require meticulous hyperparameter tuning for each task. More efficient optimization methods are needed to align with computational or temporal constraints.

%% file: section/Discussion.tex
\section{Discussion}
\label{Discussion}
In this section, we discuss some open questions and seek to explore their essence and underlying principles.
Specifically, 
we discuss what knowledge LLMs have learned in \S \ref{Learned}, 
examine the fragility of the learned knowledge in application in \S \ref{Challenges}, 
analyze the dark knowledge not yet learned by machines or humans in \S \ref{Dark Knowledge}, 
and explore how LLMs can expand the boundaries of unknown knowledge from interdisciplinary perspectives \S \ref{How to benefit from interdisciplinarity?}.

\subsection{What Knowledge Have LLMs Learned?}
\label{Learned}

\textit{Critics question whether LLMs truly have knowledge} or if they are merely mimicking \cite{DBLP:journals/corr/abs-2404-15146}, akin to the ``Stochastic Parro'' \cite{DBLP:conf/fat/BenderGMS21} and ``Clever Hans'' \cite{DBLP:conf/eacl/ShapiraLAZCGSS24}.
We first review the doubts from the following three levels through \textit{observation phenomena}:
1) Memorization: LLMs primarily rely on positional information over semantic understanding \cite{DBLP:conf/acl/LiLSDSLJJL22} to predict answers.
Additionally, LLMs may generate different answers for the same question due to different expressions.
2) Comprehension and application: \citet{DBLP:journals/corr/abs-2309-14402} argue that LLMs hardly efficiently apply knowledge from pre-training data, even when such knowledge is perfectly stored and fully extracted from LLMs.
Therefore, LLMs struggle with various reasoning tasks \cite{DBLP:journals/corr/abs-2307-02477,Alice,DBLP:journals/corr/abs-2405-14831} as well as the reversal curse \cite{DBLP:journals/corr/abs-2309-12288}. 
Besides, LLMs are not yet able to reliably act as text world simulators and encounter difficulties with planning \cite{Simulators}.
3) Creation: Although LLMs are capable of generating new terms, their quality often falls below that created by humans \cite{DBLP:journals/cacm/Raiola23}. 
Even though LLMs possess knowledge, some critics argue that current \textit{analysis methods} may only explain low-level co-occurrence patterns, not internal mechanisms.
The primary criticism asserts that the components responsible for certain types of knowledge in LLM fail to perform effectively in practical applications \cite{DBLP:conf/nips/HaseBKG23}. 
In addition, the components responsible for specific knowledge within LLMs vary under different methods. 
For these criticisms, \citet{DBLP:journals/corr/abs-2402-13731,DBLP:conf/aaai/ChenCCLZ24} propose degenerate neurons and posit that different degenerate components indeed independently express a fact.
\citet{Query} delineate the differences in the mechanisms of knowledge storage and representation, proposing the Query Localization Assumption to response these controversies.
\citet{DBLP:journals/corr/abs-2309-14316} further observe that knowledge may be memorized but not extracted due to the knowledge not being sufficiently augmented (e.g., through paraphrasing, sentence shuffling) during pretraining.
Hence, rewriting the training data to provide knowledge augmentation and incorporating more instruction fine-tuning data in the pre-training stage can effectively alleviate the above challenges and criticisms.

Despite considerable criticism, the mainstream view \cite{DBLP:journals/corr/abs-2405-12205,jinemergent,jin2024latent} is that \textbf{current LLMs may possess basic world knowledge via memorization but hardly master underlying principles for reasoning and creativity}.
In other words, LLMs master basic knowledge via memorization (discussed in \S \ref{Memorization}). 
Although LLMs possess the foundational ability to comprehend and apply knowledge (discussed in \S \ref{Comprehension and Application}), exhibiting plausible and impressive reasoning capabilities.
Current LLMs still struggle with reasoning and planning in complex tasks due to the fragility of knowledge in LLMs (elaborated in \S \ref{Challenges}). 
These reasoning and planning abilities usually require to be induced through techniques such as ICL and CoT.
Unfortunately, current LLMs are nearly incapable of creation due to the architectural limitations (discussed in \S \ref{Creation}).
Therefore, some scholars explore various architectural choices (e.g., Mamba \cite{DBLP:journals/corr/abs-2312-00752}) and training procedures.
Besides, recent research attempts to manipulate neurons, knowledge circuits, or representations \cite{DBLP:journals/corr/abs-2309-14402,RepresentationEngineering,ReFT,DBLP:conf/nips/0002PVPW23} to explore more knowledge and awaken the reasoning and planning capabilities of LLMs.

\noindent \textbf{\includegraphics[scale=0.15]{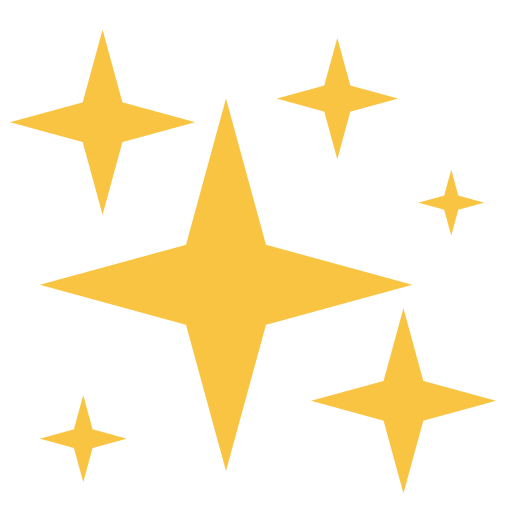} Remarks}: LLMs have learned basic knowledge of the world by \textit{momorization}.
However, the learned knowledge is fragile, leading to challenges in \textit{knowledge comprehension and application}.
Unfortunately, due to architectural limitations, current LLMs struggle with \textit{creation}.



\subsection{Why Is Learned Knowledge Fragile?}
\label{Challenges}
\textit{The knowledge learned by LLMs is fragile}, leading to challenges in application including hallucination, knowledge conflicts, failed reasoning, and safety risk \footnote{The secure risk is elaborated in \S \ref{Trustworthy LLMs}.}.
\textbf{Hallucination} denotes content generated by LLMs that diverges from real-world facts or inputs \cite{DBLP:journals/corr/abs-2311-05232,DBLP:journals/corr/abs-2401-11817,farquhar2024detecting,chen2024multi}.
On the one hand, factuality hallucination underscores the disparity between generated content and real-world knowledge.
On the other hand, faithfulness hallucination describes the departure of generated content from user instructions or input context, as well as the coherence maintained within the generated content.
\textbf{Knowledge Conflict} inherently denotes inconsistencies in knowledge \cite{DBLP:journals/corr/abs-2403-08319,DBLP:journals/corr/abs-2404-16032}.
On the one hand, internal memory conflicts within the model cause LLMs to exhibit unpredictable behaviors and generate differing results to inputs which are semantically equivalent but syntactically distinct \cite{DBLP:journals/corr/abs-2403-08319,DBLP:journals/corr/abs-2310-07521,DBLP:journals/corr/abs-2311-05876,DBLP:journals/corr/abs-2211-05853}.
On the other hand, context-memory conflict emerges when external context knowledge contradicts internal parametric knowledge \cite{DBLP:journals/corr/abs-2403-08319,DBLP:conf/acl/MallenAZDKH23}.

We posit that \textbf{these challenges mainly derive from improper learning data.}
Specifically, {hallucination} is introduced by data \cite{DBLP:conf/emnlp/KangC23,ExtrinsicHallucinations,zhang2024knowledge}, heightened during the pre-training \cite{DBLP:conf/nips/BrownMRSKDNSSAA20,DBLP:conf/acl/0001C22}, alignment \cite{DBLP:journals/corr/abs-2304-13734,DBLP:conf/nips/Ouyang0JAWMZASR22}, and deficiencies in decoding strategies \cite{DBLP:conf/acl/LewisDF18,DBLP:journals/corr/abs-2309-03883,DBLP:journals/corr/abs-2305-14739}.
Internal memory conflict can be attributed to training corpus bias \cite{DBLP:conf/emnlp/WangMWZC23}, and exacerbated by decoding strategies \cite{DBLP:conf/nips/LeePXPFSC22} and knowledge editing.
Context-memory conflict arises mainly from the absence of accurate knowledge during training, necessitating retrieval from databases and the Web.
Failed reasoning usually arises from improper \textbf{data distribution}.
Specifically, knowledge may be memorized but not extractable or applicable without sufficient augmentation (e.g., through paraphrasing, sentence shuffling) during pre-training \cite{DBLP:journals/corr/abs-2309-14316}.
\citet{GeneralizationMemorization} also delve into the mechanism between parametric knowledge and learning data, demonstrate that training data distribution qualitatively influences generalization behavior \cite{jiang2024taia}.
\citet{Grokked} further suggest that improper data distribution in the corpus causes LLMs to lack essential reasoning components, such as the bridge layer for two-hop reasoning.
Similar mechanism analysis also supports the above conclusion, indicating that hallucinations arise from a lack of mover heads \cite{knowledgecircuit,DBLP:journals/corr/abs-2403-18167}, while knowledge conflicts stem from circuit competition failure in the last few layers \cite{DBLP:journals/corr/abs-2403-19521,merullo2023mechanism,DBLP:conf/nips/HaseBKG23,DBLP:conf/coling/Ju0DYRL24,DBLP:journals/corr/abs-2402-18154}.
Additionally, \textbf{data quantity} is crucial for knowledge robustness.
Specifically, LLMs can systematically learn comprehensive understandings of the world from extensive datasets, while little data during post-training stage may compromise the robustness of knowledge representation.
This assumption is confirmed by numerous failures of post-training.
For example, SFT exacerbates hallucinations \cite{DBLP:journals/corr/abs-2405-05904,DBLP:journals/corr/abs-2403-05612}, and knowledge editing amplifies knowledge conflicts \cite{DBLP:journals/corr/abs-2310-02129,DBLP:journals/corr/abs-2402-09656}.
Note that {safety issues} usually caused by the distribution of unseen data (adversarial input) \cite{DBLP:conf/nips/0001HS23,DBLP:journals/corr/abs-2401-06824}, which is elaborated in \S \ref{Trustworthy LLMs}.

\noindent \textbf{\includegraphics[scale=0.15]{fig/remark.png} Remarks}: Improper learning caused by data distribution and quantity might be the fundamental and primary cause.


\subsection{Does Difficult-to-Learn ``Dark Knowledge'' Exist?}
\label{Dark Knowledge}

The distribution and quality of data are vital for knowledge acquisition and robust operation within the model (machine).
Imagine an ideal scenario where we have access to all kinds of data to train the machine.
The data includes all possible modalities, such as text, image, audio, video, etc.
Models can also interact with each other and the external environment.
In this long-term development, will there still be unknown dark knowledge for intelligence to human or model (machine)?

\begin{figure}[!t]
    \centering
    \includegraphics[width=0.5\textwidth]{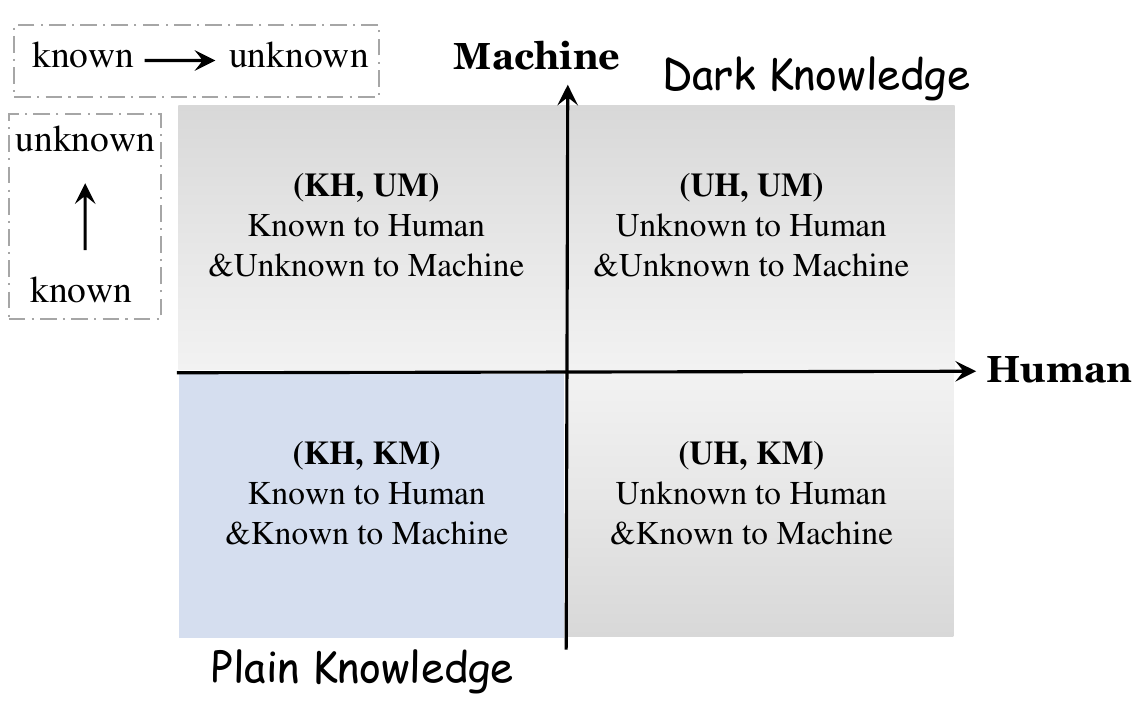}
    \caption{The future cognition of knowledge. The direction of the arrow represents the transition of knowledge from known to unknown.
    Dark knowledge, represented in gray, denotes knowledge unknown to human or machine.
    Plain knowledge known to both human and machine is highlighted in blue.
    }
    \label{fig:dark}
    \vskip -0.1in
\end{figure}

We hypothesize that there will still \textbf{exist dark knowledge for intelligence} in the future.
As shown in Fig \ref{fig:dark}, dark knowledge describes knowledge unknown to human or machine from the following three situations:
1) knowledge unknown to human \& known to machine (UH, KM).
Machines leverage vast amounts of data to explore internal patterns, whereas humans struggle with processing such data due to physiological limitations on data processing capacity and computational limits \cite{DBLP:journals/corr/abs-2312-09390,LLMCritics}. (UH, KM) includes gene prediction, intelligent transportation systems, and more.
Specifically, the structural elucidation of proteins remains mysterious to humans for a long time.
Cryo-electron microscopy, through capturing millions of images, first reveals the three-dimensional structures of proteins. 
Now, neural models can directly predict protein properties with high efficiency and accuracy \cite{pak2023using}.
2) knowledge known to human \& unknown to machine (UH, KM).
On the one hand, some scholars claim that machine can possess a ``Theory of Mind'' capability \cite{Beliefs} and emotions \cite{DBLP:conf/vr/NormoyleSD24}.
On the other hand, critics contend that machine lacks sentience \cite{DBLP:journals/socomp/AlveroP23} and merely probabilistically generates tokens.
The causes, extent, and dynamics of these emotions and sentience (like hunger, happiness, and loneliness) are subtle and intricate, making precise mathematical modeling by the machine exceptionally challenging.
Specifically, different factors are tightly coupled, making it nearly impossible to disentangle clear input-output relationships as with well-defined factual knowledge.
The sentient knowledge also exhibits chaotic behavior \cite{DBLP:journals/access/LiZY20a,DBLP:journals/complexity/DebboucheOBGKJA21}, being highly sensitive to initial conditions, where small changes can lead to vastly different outcomes \cite{DBLP:conf/complexnetworks/Segretain0TG20}.
Therefore, opponents argue that no matter how many parameters machine possesses, it cannot learn all the knowledge that human has mastered.
3) knowledge unknown to human \& unknown to machine (UH, UM) is beyond our cognition, e.g., the uncertainty in quantum mechanics and the origin of the universe.
Generally, Dark knowledge extends beyond current data and model architectures \cite{GoAIs}.
(UH, UM) necessitates human-machine collaboration.
Yet, there is no definitive conclusion on whether (UH, KM) and (KH, UM) will be solved by model architecture, training data, and computational resources.
Note that plain knowledge known to human and machine in Fig \ref{fig:dark} encompasses well-defined historical events, mathematical theorems, physical laws, etc. 

\noindent \textbf{\includegraphics[scale=0.15]{fig/remark.png} Remarks}: Dark knowledge may persist for a long time and requires human-machine collaboration to explore. 

\subsection{How to Explore More Knowledge from Interdisciplinary Inspiration?}
\label{How to benefit from interdisciplinarity?}

How can LLMs continuously narrow the boundaries of dark knowledge and achieve higher level intelligence by leveraging the human experience of perpetual knowledge exploration throughout history? 
We may draw inspirations from the following interdisciplinary studies.

\noindent \textbf{Neuroscience} studies the structure and function of the brain at molecular, cellular, neural circuit, and neural network levels \cite{squire2012fundamental}.
Generally, both mechanism analysis in LLMs and neuroscience utilize observation and intervention methods to investigate the basic principles of knowledge learning and memory, decision-making, language, perception, and consciousness.
The biological signals of the human brain and the internal activation signals in LLMs are capable of reciprocal transformation \cite{caucheteux2023evidence, feng2023aligning,TransformerDebugger,flesher2021brain}.
Benefiting from advancements in neuroscience \cite{jamali2024semantic,thiebaut2022emergent,lee2022solving}, mechanism analysis in LLMs has identified analogous function neurons and regions \cite{functionregion}, and knowledge circuits \cite{knowledgecircuit}. 
Besides, leveraging plasticity theory in neuroscience, LLMs explain the underlying technical support for intelligence \cite{DBLP:journals/corr/abs-2403-06996}.
In the future, mechanism analysis of LLMs may draw inspirations from neuroscience, guiding the next generation of artificial intelligence in organizing neural frameworks and in the storage and utilization of knowledge \cite{ren2024brain,PNN,ExplicitMemory}.

\noindent \textbf{Cognitive Science} focuses on the mind and its processes \cite{kolak2006cognitive,baronchelli2013networks}, which include language, perception, memory, attention, reasoning, emotion and mental state.
Although cognitive science and neuroscience overlap in their research content, cognitive science focuses more on abstract knowledge such as mental states and emotions rather than specific knowledge.
Therefore, \citet{Beliefs} track beliefs of self and others (formulated as ``Theory of Mind'') in LLMs from the psychological perspective within cognitive science.
\cite{DBLP:conf/iclr/WangZX022} further observe social-cognitive skill in multi-agent communication and cooperation.
Generally, there is potential to explore advanced cognitive capabilities in LLMs from the perspective of cognitive science \cite{PositionPaper}.

\noindent \textbf{Psychology} is the scientific study of mind and behavior, which include both conscious and unconscious phenomena, and mental processes such as thoughts, feelings, and motives.
Benefiting from decades of research in human psychology, machine psychology aims to uncover mechanisms of decision-making and reasoning in LLMs by treating them as participants in psychological experiments \cite{DBLP:journals/corr/abs-2303-13988}.
Machine psychology may delve into mysteries of social situations and interactions
shaping machine behavior, attitudes, and beliefs \cite{DBLP:conf/uist/ParkOCMLB23}.
Besides, group psychology paves an auspicious path for exploring dynamics such as debates and collaboration among LLMs (agents).
For instance, \textit{Dunning–Kruger effect} \cite{DBLP:conf/cogsci/MahmoodiAB13,DBLP:conf/acsos/BrownE20} in cognitive psychology filed describes that individuals with limited competence in a particular domain overestimate their abilities, and vice versa.
This phenomenon may guide the final vote in group debates and discussions.
Promisingly, psychology of learning can be applied to study prompt designs, boost learning efficiency, improve communication strategies, and develop feedback mechanisms for LLMs \cite{DBLP:journals/corr/abs-2401-10904}.

\noindent \textbf{Education} is the transmission of knowledge, skills, and character traits and manifests in various forms. 
Inspired by education in humans, \citet{DBLP:journals/corr/abs-2401-01286} categorize knowledge acquisition in LLMs into three distinct phases: recognition, association, and mastery. 
Besides, education instructs humans managing various types of conflicts: identifying inconsistencies in external information (inter-context conflict), deciding between external sources and internal memory (context-memory conflict), resolving memory confusion (internal memory conflict), and addressing cultural conflicts.
The above knowledge conflicts and integration also exist in knowledge evolution of LLMs across individuals and groups \cite{DBLP:journals/corr/abs-2308-02773}.
Fortunately, education facilitates humans in learning to learn.
Can LLMs similarly self-evolve to continuously adapt to societal changes and requirements?

\noindent \textbf{\includegraphics[scale=0.15]{fig/remark.png} Remarks}:
LLMs may improve their architecture and mechanisms for knowledge learning, storage, and expression, drawing inspiration from neuroscience. 
Besides, cognitive science and psychology provide promising alternatives for sophisticated intelligence, emergent capabilities and behaviors in evolution.
Educational studies can inspire the learning strategy of LLMs, navigating conflicts and integrating knowledge during their evolution.



%% file: section/Future_Directions.tex
\section{Future Directions}
\label{Future_Directions}

\subsection{Parametric VS. Non-Parametric Knowledge}
LLMs can be conceptualized as parametric knowledge stores, where the parameters of the model—typically the weights of the neural network—encode a representation of the world’s knowledge. 
This parametric approach to knowledge storage means that the knowledge is implicitly embedded within the model’s architecture, and it can be retrieved and manipulated through the computational processes of the neural network \cite{DBLP:journals/corr/abs-2309-14402}.
In contrast, non-parametric knowledge storage involves methods where the knowledge is explicitly represented and can be directly accessed. Examples of non-parametric knowledge storage include knowledge graphs, databases, and symbolic reasoning systems, where knowledge is represented as discrete symbols or facts.
Parametric knowledge enables LLMs to deeply compress and integrate information \cite{DBLP:journals/corr/abs-2404-09937,DBLP:journals/entropy/ShwartzZivL24}, allowing them to generalize and apply this knowledge across various contexts. This is akin to LLMs mastering the mathematical operation rule of ``mod'' through parametric knowledge, enabling them to generalize and seamlessly solve all mod-related problems \cite{MemorizeorGeneralize, DBLP:journals/corr/abs-2402-17709}.
Conversely, non-parametric knowledge requires extensive searches across the knowledge space for each user query.
Subsequently, \citet{Grokked} also prove that non-parametric knowledge severely fails in complex reasoning tasks, with accuracy levels approaching random guessing.
Unfortunately, parametric knowledge within LLMs is opaque, often encountering challenges such as interpretability issues, outdated information, hallucinations, and security concerns.

Addressing these issues often requires leveraging external non-parametric knowledge, which offers transparency, flexibility, adaptability, and ease of operation.
However, \textit{augmenting parametric knowledge} in LLMs with non-parametric knowledge \cite{ExplicitMemory,DBLP:journals/corr/abs-2310-01061,DBLP:journals/corr/abs-2308-09729,ko2024investigating} remains an ongoing challenge due to retrieval accuracy from haystack, context lengths, and resources \footnote{On the one hand, storing large amounts of non-parametric knowledge requires a lot of space and high maintenance costs. On the other hand, retrieving information for each user query is very resource-intensive.} limitations \cite{shang2024ai,DBLP:journals/corr/abs-2402-19473}.
Besides, simultaneously retrieving relevant information from a long context and conducting reasoning is nearly impossible in reasoning-in-a-haystack experiments \cite{shang2024ai}.
Similarly, \textit{augmenting non-parametric knowledge}—either by distilling knowledge from an LLM's parametric knowledge \cite{DBLP:conf/naacl/WestBHHJBLWC22,DBLP:journals/corr/abs-2301-11293} or by using it to parse text directly \cite{DBLP:conf/acl/ZhangG023}—also poses significant challenges.
Moreover, \citet{ExplicitMemory} propose \textit{a novel explicit memory that lies between parametric and non-parametric knowledge}. 
LLM with explicit memory enjoys a smaller parameter size and lower resource consumption for retrieving external non-parametric knowledge.

Generally, \textbf{inspired by the knowledge mechanisms analysis in LLMs, we have the potential to develop more architectural and learning strategies for organizing knowledge within LLMs}.
These efficient LLMs \cite{DBLP:journals/corr/abs-2402-08797} are advancing toward lower GPU, computation, and storage resource requirements, as well as smaller model sizes by combining the strengths of parametric and non-parametric knowledge \cite{ExplicitMemory,PNN,Chen_2024,Pan_2024,pan2023largelanguagemodelsknowledge}.

\subsection{Embodied Intelligence}
\label{Embodied Intelligence}
The current LLM still cannot be regarded as a truly intelligent creature~\cite{bender-koller-2020-climbing,bisk-etal-2020-experience}. 
The process of human language acquisition is not merely a passive process of listening to language.
Instead, it is an active and interactive process that involves engagement with the physical world and communication with other people. 
To enhance the current LLM's capabilities and transform it into a powerful agent, it is necessary to enable it to learn from multimodal information and interact with the environment and humans.
\paragraph{Multimodal LLMs.}
The integration of multiple modalities is a critical challenge in the field of LLMs and embodied AI.
While LLMs have demonstrated impressive capabilities when processing language data, their ability to seamlessly incorporate and synthesize information from other modalities such as images, speech, and video is still an area of active research.
However, the current multi-modal model faces challenges, particularly in complex reasoning tasks that require understanding and integrating information from both text and images.

Recent studies~\cite{huang2024vlkeblargevisionlanguagemodel,DBLP:journals/corr/abs-2403-20330} have highlighted the discrepancy between the model’s performance in language tasks and its ability to integrate knowledge from different modalities effectively. 
These findings suggest that current models often prioritize linguistic information, failing to fully exploit the synergistic potential of multimodal data \cite{DBLP:journals/corr/abs-2402-07233}.
There are some pioneering efforts in this direction ~\cite{pan2024findingeditingmultimodalneurons,DBLP:conf/iccvw/SchwettmannCKB023} , aiming to uncover the mechanisms by which multi-modal models store and retrieve information.
Despite these advancements, there is still a need for further exploration to deepen our understanding of multi-modal knowledge storage. 

\paragraph{Self-evolution.}
As discussed in the previous part, current language models are mainly based on tuning to gain knowledge, which requires a lot of training and high-quality data.
These learnings are passive whereas, to be a human, evolution usually also undergoes communication and interaction.
As an intelligent agent, the models should be able to learn through interactions and learn by themselves spontaneously.
Recently, some work has attempted to enable the model to learn by themselves~\cite{zhang2024selftuninginstructingllmseffectively} or learn by interaction with the environment~\cite{xu2024interactiveevolutionneuralsymbolicselftraining,xi2024agentgymevolvinglargelanguage}.
By integrating self-evolving mechanisms, models can continuously update their knowledge base and improve their understanding without relying solely on manually curated datasets.
This not only reduces the dependency on large-scale labeled data but also allows the models to adapt to evolving linguistic norms and cultural contexts over time.

%

\subsection{Domain LLMs}
\label{appendix:Domain LLMs}
The success of general-purpose LLMs has indeed inspired the development of domain-specific models that are tailored to particular areas of knowledge \cite{Stakeholders}, such as 
biomedicine~\cite{yu2024largelanguagemodelsbiomedical,DBLP:journals/corr/abs-2405-01469}, finance~\cite{yang2023fingptopensourcefinanciallarge}, geoscience~\cite{deng2023k2foundationlanguagemodel},
ocean science~\cite{bi2024oceangptlargelanguagemodel}, etc.
However, unlike human language, the knowledge of these different domains bears specific characteristics.
It remains unclear whether LLMs can acquire complex scientific knowledge or if such knowledge still resides within the realm of current dark knowledge.
Furthermore, does domain-specific knowledge such as mathematics share the same underlying mechanisms as textual knowledge \cite{DBLP:journals/corr/abs-2403-04571}, or does it exhibit more intricate mechanisms of knowledge acquisition?
Currently, there is a relative lack of research focusing on the mechanism of these domain-specific knowledge and there is an increasing recognition of the importance of developing a deeper understanding of these mechanisms. 

Data sparsity and diversity in domain-specific models pose another challenge.
Sparsity is usually caused by confidentiality, privacy, and the cost of acquisition in specialized fields.
As for diversity, the presentation of knowledge varies across different fields.
For instance, in the biomedical domain, knowledge includes complex biological concepts such as the structure and function of proteins and molecules. This requires models to integrate understanding that extends beyond natural language, often involving graphical representations like chemical structures, which cannot be directly expressed in text. 
Similarly, in fields such as finance and law~\cite{lai2023largelanguagemodelslaw}, models must engage in sophisticated reasoning and decision-making processes based on domain-specific knowledge.
Hence, the critical tasks of collecting high-quality data for domain-specific models (including synthetic data generation) and effectively embedding domain knowledge into LLMs require immediate attention.


%% file: section/Conclusion.tex
\section{Conclusion}
\label{Conclusion}
In this paper, we propose a novel knowledge mechanism analysis taxonomy and review knowledge evolution. 
We further discuss knowledge utilization issues, as well as unexplored dark knowledge.
We hope these insights may inspire some promising directions for future research and shed light on more powerful and trustworthy models.



%% file: section/appendix.tex
\appendix

\section{Comparison of Methods for Knowledge Evolution}
\label{appendix:Comparison of Methods for Knowledge Evolution}

\begin{figure}[!t]
    \centering
    \includegraphics[width=0.5\textwidth]{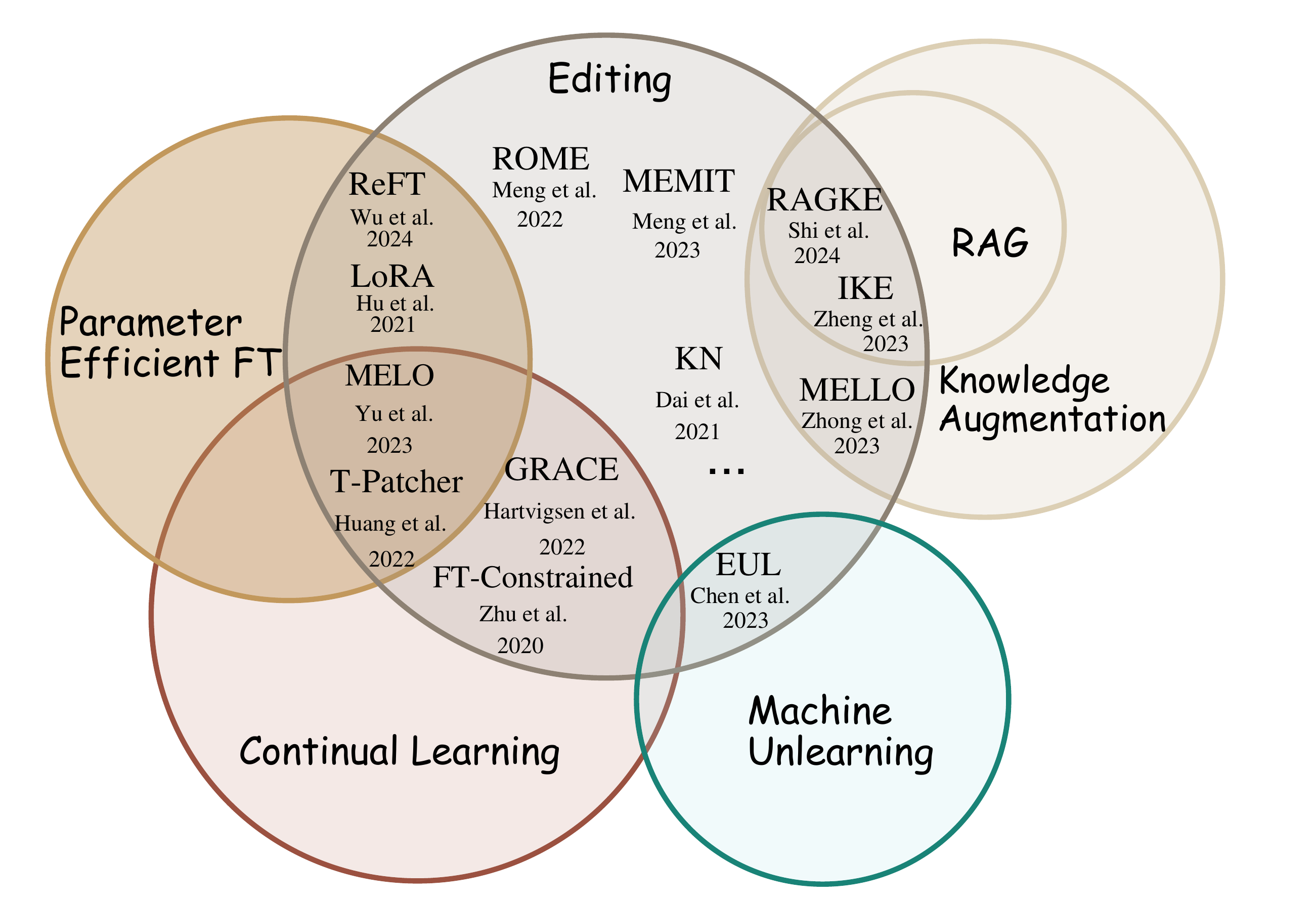}
    \caption{Comparison of Different Methods for Knowledge Evolution.}
    \label{fig:comparison}
    \vskip -0.15in
\end{figure}

Note that due to the page limit, \S \ref{Knowledge Evolution} does not provide a detailed enumeration of various techniques and details, such as machine unlearning and knowledge augmentation.
Hence, we briefly outline common methods during post-training stage in this section and illustrate their associations and differences \cite{DBLP:journals/corr/abs-2401-01286} in Fig \ref{fig:comparison}.

\begin{itemize}[leftmargin=*,nosep]
    \item \textit{Continual Learning} aims to continually acquire new skills and learn new tasks while retaining previously acquired knowledge.
    \item \textit{Parameter-efficient Fine-tuning} (PET) \cite{DBLP:conf/acl/ZhangHLJSL19} only updates a minimal set of parameters instead of full fine-tuning. A promising strategy is LoRA \cite{DBLP:conf/iclr/HuSWALWWC22}.
    \item \textit{Knowledge Augmentation} is proposed to assist the model in handling unknown knowledge for LLMs \cite{DBLP:conf/acl/ZhangHLJSL19,DBLP:journals/aiopen/HanZDLS22}. RAG \cite{DBLP:journals/corr/abs-2404-10981} is the is the most prevalent methods. Beside, knowledge augmentation also includes prompt engineering \cite{DBLP:journals/corr/abs-2307-12980,DBLP:journals/interactions/KraljicL24,DBLP:journals/corr/abs-2305-19118} and in-context learning \cite{DBLP:journals/corr/abs-2401-11624}.
    \item \textit{Machine Unlearning} \cite{nguyen2022survey,tian2024forget,DBLP:journals/corr/abs-2403-13682} focuses on discarding undesirable behaviors from LLMs.
    \item \textit{Editing}, including knowledge editing \cite{DBLP:journals/corr/abs-2401-01286} and representation editing \cite{ReFT}, aims to enable quick and precise modifications to the LLMs. 
    Usually, editing first identifies the knowledge location in LLMs and then precisely modifies model behavior through a few instances. 
\end{itemize}

\section{Universality Intelligence}
\label{appendix:Universality Intelligence}

To validate the hypotheses in this paper across different architectures, we first introduce other popular model architectures in \S \ref{appendix:Model Architecture}, and observe the generalizability of our hypotheses across other model architectures in \S \ref{appendix:Knowledge Mechanisms in other Architectures}.
Besides, recent work further claims that models trained with different data, modalities, and objectives are converging to shared representation spaces \cite{huh2024platonic}. 
Artificial and biological neural networks \footnote{Unless otherwise specified as biological networks, the terms models, neural networks, neural models, and machines refer to artificial neural networks.} also share similar features and circuits, suggesting a universal underlying mechanism \cite{DBLP:journals/corr/abs-2310-13018,NaturalAbstractions,DBLP:conf/icml/Kornblith0LH19}.
Therefore, analogous to biological taxonomy, we introduce artificial neural model family and discuss the potential universality intelligence in the future in \S \ref{appendix:Machine and Human}.



\subsection{Model Architecture}
\label{appendix:Model Architecture}

\subsubsection{Transformer}

\paragraph{MLP}
The Multilayer Perceptron (MLP) is a crucial component in neural networks, usually comprising multiple fully connected layers. Within the Transformer architecture, the MLP plays a vital role in applying nonlinear transformations to the input hidden states, thereby enriching the model's capacity for expression. More precisely, every MLP block involves two linear transformations separated by a point-wise activation function $\sigma$:
\begin{equation}
\begin{aligned}
  \text{MLP}^l\left( h^l \right) = \sigma\left( W_K^l h^l \right) W_V^l,
\end{aligned}
\label{transformer-eq}
\end{equation}
where $\sigma$ is the point-wise activation function, typically a non-linear function such as ReLU or GELU. $W_K^l$ is the weight matrix for the first linear transformation in the $l$-th layer, mapping the input hidden state $h^l$ to an intermediate representation. $W_V^l$ is the weight matrix for the second linear transformation in the $l$-th layer, transforming the intermediate representation to the output of the MLP block.

\paragraph{Attention}is a mechanism in neural networks, especially in models like Transformers, that captures dependencies between different positions within a sequence. It works by transforming each input element into Query ($Q$), Key ($K$), and Value ($V$) vectors, computing attention scores between elements, and then calculating a weighted sum of values based on these scores. Specifically, for an input sequence represented as matrix $X$, the transformations are as follows:
\begin{equation}
\begin{aligned}
  Q &= XW^Q, \\
  K &= XW^K, \\
  V &= XW^V,
\end{aligned}
\end{equation}
where $W^Q$, $W^K$, and $W^V$ are learned projection matrices. The attention scores are computed using the scaled dot-product attention mechanism:
\begin{equation}
\begin{aligned}
  H = \text{Attention}(Q, K, V) = \text{Softmax}\left( \frac{QK^T}{\sqrt{d_k}} \right)V,
\end{aligned}
\end{equation}
where $d_k$ is the dimensionality of the Key vectors. This allows the model to focus on different parts of the sequence adaptively, making it effective for tasks like natural language processing where understanding long-range dependencies is important.

\paragraph{Variants of Transformer}
Variants of the Transformer also achieve success.
For instance, RWKV \cite{DBLP:conf/emnlp/PengAAAABCCCDDG23} combines the efficient parallelizable training of transformers with the efficient inference of RNNs while mitigating their limitations.
TTT \cite{TTT} replaces the hidden state of an RNN with a machine learning model. TTT compresses context through actual gradient descent on input tokens.
RetNet \cite{DBLP:journals/corr/abs-2307-08621} theoretically derives the connection
between recurrence and attention, simultaneously achieves training parallelism,
low-cost inference, and good performance.

\subsubsection{SSM}
\paragraph{Mamba}introduced by \citet{DBLP:journals/corr/abs-2312-00752}, is a recent family of autoregressive language models based on state space models (SSMs). Mamba employs a unique architecture called MambaBlock, which replaces the attention and MLP blocks used in Transformer layers.

Specifically, Mamba maps a sequence of tokens $x = [x_1, x_2, \ldots, x_T]$ to a probability distribution over the next token $y$. Each token $x_i$ is first embedded into a hidden state of size $d$ as $h_i^{(0)}$, which is then transformed sequentially by a series of MambaBlocks. The hidden state $h_i^{(\ell)}$ after the $\ell$-th MambaBlock is computed as follows:

\begin{equation}
\begin{aligned}
h_i^{(\ell)} = h_i^{(\ell-1)} + o_i^{(\ell)}
\end{aligned}
\end{equation}

The output $o_i^{(\ell)}$ of the $\ell$-th MambaBlock for the $i$-th token is a combination of $s_i^{(\ell)}$ (from Conv and SSM operations) and $g_i^{(\ell)}$ (a gating mechanism):

\begin{equation}
\begin{aligned}
o_i^{(\ell)} &= \text{MambaBlock}^{(\ell)} \left[h_1^{(\ell-1)}, h_2^{(\ell-1)}, \ldots, h_i^{(\ell-1)}\right]  \\
&= W_o^{(\ell)} \left[ s_i^{(\ell)} \otimes g_i^{(\ell)} \right]
\end{aligned}
\end{equation}

Here, $\otimes$ denotes element-wise multiplication. The calculation of $s_i^{(\ell)}$ is as follows:

\begin{equation}
\begin{aligned}
a_i^{(\ell)} = W_a^{(\ell)} h_i^{(\ell)}
\end{aligned}
\end{equation}

\begin{equation}
\label{eq:Conv}
\begin{aligned}
&c_1^{(\ell)}, c_2^{(\ell)}, \ldots, c_i^{(\ell)} = \\
&\text{SiLU} \left[ \text{Conv1D} \left[ a_1^{(\ell)}, a_2^{(\ell)}, \ldots, a_i^{(\ell)} \right] \right]
\end{aligned}
\end{equation}

\begin{equation}
\label{eq:SSM}
\begin{aligned}
s_i^{(\ell)} = \text{selective-SSM} \left[ c_1^{(\ell)}, c_2^{(\ell)}, \ldots, c_i^{(\ell)} \right]
\end{aligned}
\end{equation}

The operations in Equations (\ref{eq:Conv}) and (\ref{eq:SSM}) correspond to Conv and SSM operations, respectively. The gating mechanism $g_i^{(\ell)}$ is given by:

\begin{equation}
\begin{aligned}
g_i^{(\ell)} = \text{SiLU} \left[ W_g^{(\ell)} h_i^{(\ell-1)} \right]
\end{aligned}
\end{equation}

The formulas and concepts used here are adapted from \citet{DBLP:journals/corr/abs-2404-03646}.

Compared to Transformer, Mamba's design enables more efficient parallel training and effectively captures dependencies in sequences, making it suitable for various natural language processing tasks.

\subsubsection{Vision and Multi-modal Models}
\label{appendix:Vision Model}

In the realm of vision and multi-modal models, various architectures have emerged, each with its unique approach to tackling complex visual tasks. For example, \textbf{GANs} (Generative Adversarial Nets) \cite{DBLP:conf/nips/GoodfellowPMXWOCB14} consist of two neural networks: a generator and a discriminator. Through adversarial learning, the generator aims to produce realistic data samples (such as images), while the discriminator attempts to distinguish between real and generated data. \textbf{Diffusion Model} \cite{DBLP:journals/corr/abs-2308-09388,DBLP:conf/icml/Sohl-DicksteinW15} is a powerful tool for generating high-quality images and data. It simulates a diffusion process by gradually adding and removing noise to achieve data generation. \textbf{ResNet} (Residual Network) \cite{DBLP:conf/cvpr/HeZRS16} introduced residual learning, revolutionizing deep network training by improving efficiency and performance through skip connections. 
\textbf{ViT} (Vision Transformer) \cite{DBLP:conf/iclr/DosovitskiyB0WZ21} integrated the Transformer architecture into vision tasks, capturing long-range dependencies by processing image patches.

\subsection{Knowledge Mechanisms in Other Architectures}
\label{appendix:Knowledge Mechanisms in other Architectures}

Surprisingly, similar mechanisms as those found in transformer-based LLMs have also been discovered in other architectural models. 
Specifically, Mamba employs the knowledge memorization mechanism similar to Transformer \cite{DBLP:journals/corr/abs-2404-03646}.
Vision and multi-modal architectures also adopt function region (\textit{Modular Region Hypothesis}) for knowledge utilization \cite{DBLP:journals/corr/abs-2311-07470,DBLP:conf/iccvw/SchwettmannCKB023,DBLP:conf/icml/KohNTMPKL20,DBLP:conf/cvpr/BauZKO017}, e.g., multi-modal neuron regions are responsible for multi-modal tasks.
Besides, the \textit{connections hypothesis} between neurons is found in vision architecture models \cite{motif}.
\citet{motif} further suggest that different types of knowledge reuse partial components, e.g., cars and cats reuse the same neurons (\textit{Reuse Hypothesis}).
As for the Dynamic Intelligence Hypothesis, it inherently focuses on entire artificial neural models.
Generally, neural models across various architectures, trained with different objectives
on different data and modalities, are converging to a
shared statistical model of reality in their representation spaces \cite{huh2024platonic}.
These neural models may tend to share similar knowledge mechanisms and imagination \cite{Imagination}


\subsection{Machine and Human}
\label{appendix:Machine and Human}

Analogous to the Hominidae Family in biological taxonomy, artificial neural models can be regarded as Neural Model Family:
\begin{itemize}[itemsep=2pt, topsep=2pt]
    \item Family: {Neural Model}, likened to ``Hominidae''.
    \item Genus: Transformer architecture, Mamba architecture, etc., likened to ``Homo'' and ``Pan''.
    \item Species: BERT, GPT, Llama, Mistral, Mamba, etc., likened to ``Sapiens'', ``Pan troglodytes'', and ``Pan paniscus''.
\end{itemize}
Metaphorically, Llama-7B, Llama-13B, Llama-70B, etc., can be viewed as the infancy, childhood, and adulthood of humans.
\citet{shah2024development} further find that, regardless of model size, the developmental trajectories of PLMs consistently exhibit a window of maximal alignment with human cognitive development.
Therefore, we hypothesize that \textbf{artificial neural networks (machine) and biological neural networks (human) tend to converge to universality intelligence.}
In other words, huamn and machine share similar features and circuits.

Specifically, extensive evidences demonstrate that machine and human share the same mechanism of knowledge memorization, i.e., modular region and connection \cite{thiebaut2022emergent}.
The activations of modern language models can also linearly map onto the brain responses to speech \cite{caucheteux2023evidence}.
\citet{caucheteux2023evidence} pioneer the explanation via predictive coding theory: while transformer-based LLMs are optimized to predict nearby words, the human brain would continuously predict a hierarchy of representations that spans multiple timescales.
The above phenomenon indicates that machine and human share similar underlying mechanisms of knowledge \cite{DBLP:journals/corr/abs-2310-13018,NaturalAbstractions,DBLP:conf/icml/Kornblith0LH19}, irrespective of their specific configurations, process and comprehend information. 
This could be due to inbuilt inductive biases \cite{DBLP:journals/corr/abs-2403-06996} in neural networks or natural abstractions \cite{NaturalAbstractions} – concepts favored by the natural world that any cognitive system would naturally gravitate towards \cite{Review}.

\section{Tools for Mechanism Analysis}
\label{appendix:Tools}
Numerous tools exist for interpreting knowledge mechanisms in LLMs.
\textit{TransformerLens} \cite{nanda2022transformerlens} is a library for the mechanistic interpretability using observation and intervention. TransformerLens allows users to cache, remove, or replace internal activations during model running.
\textit{XMD} \cite{DBLP:conf/acl/LeeKJCLN0MSP023} provides various forms of feedback via an intuitiveness, which enable explanations align with the user feedback.
\textit{NeuroX}~\cite{NeuroX} implements various interpretation methods under a unified API then provides interpretability of LLMs. 
\textit{PatchScope} \cite{ghandeharioun2024patchscopes} is a tool developed by Google that employs a novel model to elucidate the hidden states in the original model.
\textit{Transformer Debugger}~\cite{TransformerDebugger}, an interpretability tool from OpenAI, utilizes GPT-4 and sparse auto-encoders to explain language neurons.
Sparse autoencoders \cite{gao2024scaling} 
leverages sparse auto-encoders to extract interpretable features from a language model by reconstructing activations from a sparse bottleneck layer.
\textit{Transcoders} \cite{Transcoders} decomposes model computations involving MLPs into interpretable circuits.